%% file: main.tex
\documentclass[10pt,twocolumn,letterpaper]{article}

\usepackage[pagenumbers]{wacv} 
\usepackage{graphicx}
\usepackage{amsmath}
\usepackage{amssymb}
\usepackage{booktabs}
\usepackage{multirow}
\usepackage{pifont}

\newcommand{\xmark}{\ding{55}} 


\definecolor{wacvblue}{rgb}{0.21,0.49,0.74}
\usepackage[pagebackref,breaklinks,colorlinks,allcolors=wacvblue]{hyperref}

\def\wacvPaperID{2668} 
\def\confName{WACV}
\def\confYear{2026}

\newcommand{\boldmethod}{\textrm{\textbf{TextGuider}}}

\title{TextGuider: Training-Free Guidance for Text Rendering via Attention Alignment}

\author{Kanghyun Baek$^{1}$~~~~~~~~~~Sangyub Lee$^{1}$~~~~~~~~~~Jin Young Choi$^{1}$~~~~~~~~~~Jaewoo Song$^{2,3}$\\
Daemin Park$^{2}$~~~~~~~~~~Jooyoung Choi$^{2}$~~~~~~~~~~Chaehun Shin$^{2}$~~~~~~~~~~Bohyung Han$^{1,2,}$\footnotemark[1]~~~~~~~~~~Sungroh Yoon$^{1,2,4,}$\footnotemark[1]\\
$^1$Interdisciplinary Program in Artificial Intelligence, Seoul National University\\
$^2$Department of Electrical and Computer Engineering, Seoul National University\\
$^3$Global Technology Research, Samsung Electronics\\
$^4$AIIS, ASRI, INMC, ISRC, Seoul National University\\
{\tt\small \{qor6271, nickyub, jychoi999, woo.song, eoalsqkr12, jy\_choi, chaehuny, bhhan, sryoon\}@snu.ac.kr}
}

\begin{document}
\maketitle
\input{sec/0_abstract}
\input{sec/1_intro}
\input{sec/2_related_work}
\input{sec/3_method}
\input{sec/4_experiment}

\input{sec/5_conclusion}

{
    \small
    \bibliographystyle{ieeenat_fullname}
    \bibliography{main}
}

\input{sec/a_appendix}

\end{document}

%% file: sec/0_abstract.tex
\renewcommand{\thefootnote}{\fnsymbol{footnote}}
\footnotetext[1]{Correspondence to: Sungroh Yoon (sryoon@snu.ac.kr), Bohyung Han (bhhan@snu.ac.kr)}

\begin{figure*}[t!]
    \centering
    \includegraphics[width=\linewidth]{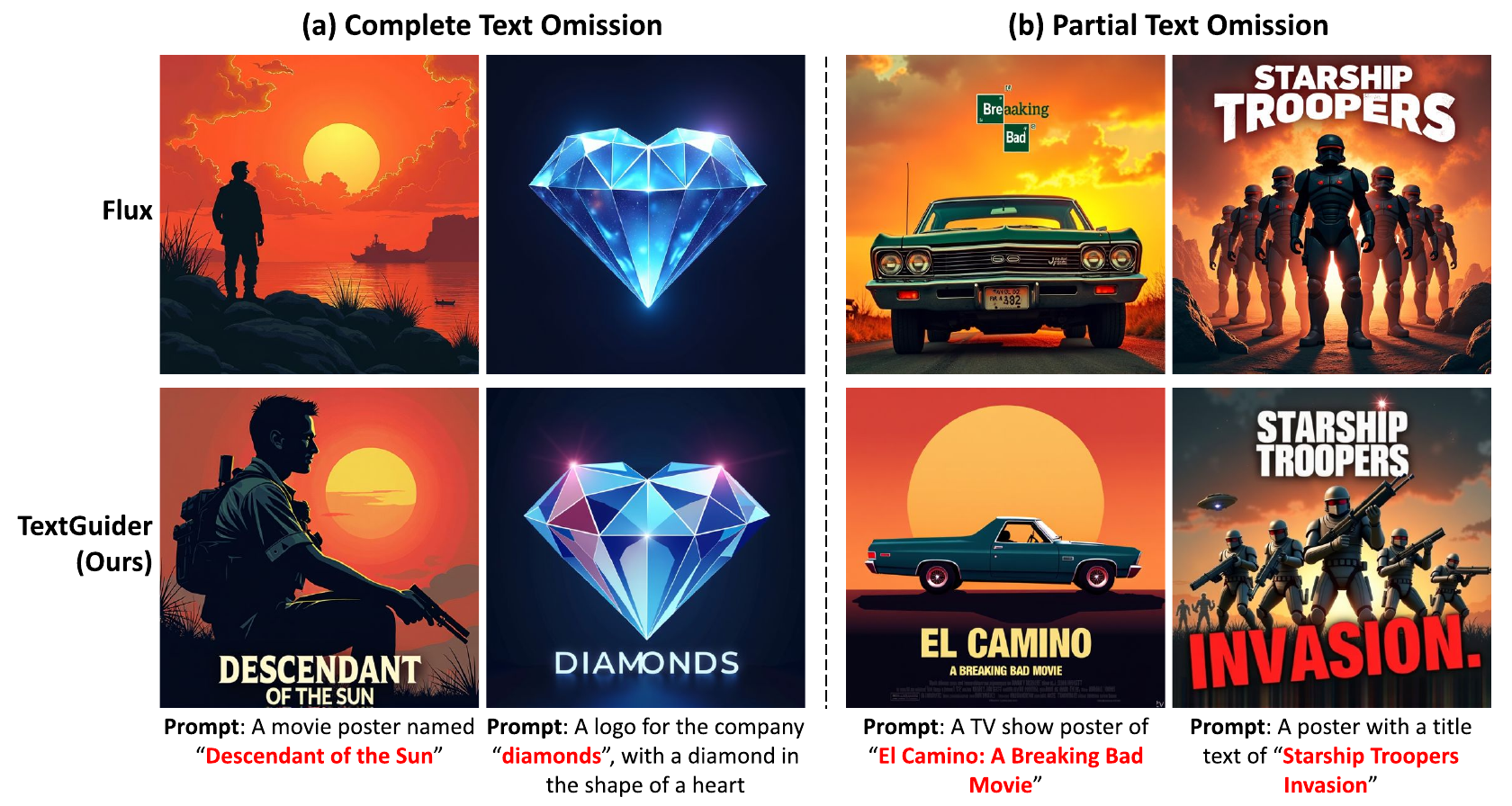}
    \caption{
    \textbf{Examples of Complete and Partial Text Omission.} (a) Complete Text Omission: The textual content is entirely missing and is represented visually instead. (b) Partial Text Omission: Specified text is only partially rendered. These issues are evident in the images generated by Flux. We investigate the text omission problem and propose a method to effectively address and mitigate it.
    }
    \label{fig:thumbnail}
\end{figure*}

\begin{abstract}
Despite recent advances, diffusion-based text-to-image models still struggle with accurate text rendering. Several studies have proposed fine-tuning or training-free refinement methods for accurate text rendering. However, the critical issue of text omission, where the desired text is partially or entirely missing, remains largely overlooked.
In this work, we propose \boldmethod, a novel training-free method that encourages accurate and complete text appearance by aligning textual content tokens and text regions in the image. Specifically, we analyze attention patterns in Multi-Modal Diffusion Transformer(MM-DiT) models, particularly for text-related tokens intended to be rendered in the image. Leveraging this observation, we apply latent guidance during the early stage of denoising steps based on two loss functions that we introduce. Our method achieves state-of-the-art performance in test-time text rendering, with significant gains in recall and strong results in OCR accuracy and CLIP score.

\end{abstract}

%% file: sec/1_intro.tex
\section{Introduction}
\label{sec:intro}

With recent advances in text-to-image (T2I) models~\cite{sohl2015deep, ddpm, dhariwal2021diffusion, ldm, sd3, peebles2023scalablediffusionmodelstransformers, flowmatching, rectifiedflow, flux1-dev}, it has become increasingly possible to generate high-quality images while enabling intuitive image control through text prompts. Despite these advancements, T2I models still struggle to handle certain aspects of real-world image generation, most notably accurate text rendering, which is crucial for many practical applications such as posters, logos, or advertisements. Several prior studies~\cite{liu2024glyph, liu2024glyph2, zhao2023udifftext, zeng2024textctrl, wang2024high} have attempted to address the limitations of text rendering by fine-tuning T2I models and text encoders on text-centric datasets~\cite{yang2023glyphcontrol, chen2023textdiffuser, chen2024textdiffuser}. However, these approaches are computationally intensive and require significant effort in dataset collection. 

More recently, the introduction of Multi-Modal Diffusion Transformer (MM-DiT) models, such as Flux~\cite{flux1-dev} and SD3~\cite{sd3}, has enabled training-free methods~\cite{hu2025amosamplerenhancingtext, du2025textcrafter} that exploit their large-scale pretraining and scalability. For instance, the Attention Modulated Overshooting (AMO) Sampler~\cite{hu2025amosamplerenhancingtext} aims to enhance text rendering by refining text regions, identified through attention maps, using an overshooting mechanism. However, the AMO Sampler becomes less effective when the attention map is not spatially aligned with regions where the text is expected to appear. Such misalignment often results in the critical yet previously underexplored problem of text omission.

Specifically, text omission refers to situations in which the intended textual content in the prompt does not properly appear in generated images, either appearing incomplete or entirely missing. For example, as shown in Figure~\ref{fig:thumbnail}, despite the Flux's ability to generate plausible visual elements, it often fails to render some or all of the intended textual content, leading to degradation in text rendering.

To analyze text omission, we compare attention maps from successful and failed generations using the same prompt (Figure~\ref{fig:success_fail_attn_map}), and observe a spatial alignment pattern that distinguishes between the two cases, particularly during the early denoising timesteps.
In successful cases, we observe that the attention map of the quotation mark token (i.e., the opening quote preceding the text) broadly covers the entire text region, while each textual content token (i.e., the text to be rendered, within quotation marks in the prompt) tends to focus on its own localized area, which assists in rendering all tokens.
In contrast, failed generations often show weak or misaligned attention in both types of tokens, which may lead to text omission in the image.


Based on this observation, we propose \boldmethod, a training-free latent guidance strategy that reinforces the alignment of the attention map across the quotation mark token, the textual content tokens, and the image regions during the early timesteps.
In particular, we introduce two novel loss functions to mitigate text omission:
a \textit{split loss}, which encourages the attention maps of individual textual content tokens to be spatially separated, and a \textit{wrap loss}, which ensures that the attention map of the quotation mark token broadly covers the regions attended to by the textual content tokens.
By applying these losses in a training-free latent guidance framework, we effectively render the desired text in the image without any model fine-tuning.

We evaluate our method and achieve state-of-the-art performance in test-time text rendering.
Notably, our approach yields a substantial improvement in recall, indicating our strategy is particularly effective at preventing text omission.
In addition, it achieves superior performance in other OCR-based metrics, as well as prompt-image alignment.
These results highlight the general applicability of our method to real-world scenarios where text rendering is critical.

Our main contributions are summarized as follows:
\begin{itemize}
\item We find that proper attention map alignment of the textual tokens during early denoising timesteps is crucial for the emergence of the text.
\item We propose a training-free latent guidance method that leverages our novel attention-based loss functions to mitigate text omission.
\item We achieve state-of-the-art performance in test-time text rendering, ensuring more complete text generation.
\end{itemize}

%% file: sec/2_related_work.tex
\section{Related Work}
\label{gen_inst}

\subsection{Scene Text Generation}

Advancements in T2I generation models~\cite{ldm,flux1-dev} have led to various methods to address the challenge of visual text rendering. Several methods~\cite{liu2024glyph, liu2024glyph2, zhao2023udifftext, zeng2024textctrl, wang2024high} leverage character-level text encoders~\cite{liu2022character, xue2022byt5} to extract glyph representations that guide the generation of the target text. In parallel, many studies incorporate additional modules or auxiliary models to explicitly control the visual appearance and placement of text. TextMastero~\cite{wang2024textmastero}, AnyText~\cite{tuo2023anytext, tuo2024anytext2}, and GlyphControl~\cite{yang2023glyphcontrol} introduce specialized glyph-aware conditioning modules, while TextDiffuser~\cite{chen2023textdiffuser, chen2024textdiffuser} employs a transformer to generate segmentation masks for layout initialization. However, all of the above methods rely on training or fine-tuning, which requires significant computational resources and risks introducing biases from the training data.

More recently, MM-DiT models~\cite{sd3, flux1-dev} have shown promising text rendering performance even without additional training, enabled by their strong text encoders and attention mechanisms. Building on this, several training-free approaches have been proposed to further improve visual text rendering. 
AMO Sampler~\cite{hu2025amosamplerenhancingtext} improves text rendering quality by applying stochastic sampling within the text regions identified by the attention map.
TextCrafter~\cite{du2025textcrafter} improves multi-text rendering by initializing separate layouts for each text instance using early attention cues. In contrast to previous methods, our approach focuses exclusively on addressing the problem of text omission.

\subsection{Training-Free Guidance}

Diffusion and flow-based generative models offer the flexibility to manipulate the sampling process at inference time by incorporating various guidance strategies. At time $t$, the latent variable $\mathbf{Z}_{t}$ is updated by a user-specified guidance function $\psi$ (\ie, $\mathbf{Z}_{t}'=\psi(\mathbf{Z}_{t})$) to steer the generation process in the desired direction reliably.
Prior work has explored various test-time guidance strategies to improve generation quality or controllability, such as classifier-based gradients~\cite{dhariwal2021diffusion}, classifier-free methods~\cite{cfg}, gradient corrections~\cite{chung2022improving}, layout constraints~\cite{chen2024training}, and self-guided control via internal representations~\cite{selfguidance}.

Several approaches~\cite{attendandexcite, a_star, syngen, zhang2024object, li2023divide, meral2024conform} have aligned complex prompts with visual content by guiding attention maps at test-time. Attend-and-Excite~\cite{attendandexcite} maximizes activations for under-attended concepts, A-STAR~\cite{a_star} reduces attention overlap and decay via dedicated losses, and SynGen~\cite{syngen} guides gradients to enhance entity-attribute alignment while suppressing interference between unrelated entities. Although test-time guidance has been actively explored for various goals, its application to the scene text generation still remains rare and underexplored.

%% file: sec/3_method.tex
\section{Method}
\label{sec:method}

\subsection{Preliminaries}

\paragraph{MM‑DiT}

MM-DiT~\cite{sd3,flux1-dev} extends the transformer-based diffusion model~\cite{peebles2023scalablediffusionmodelstransformers} to handle both image and text tokens. When rendering text using MM-DiT~\cite{hu2025amosamplerenhancingtext, du2025textcrafter}, prompts typically include quotation marks to explicitly specify the text to be rendered. The quoted content is then tokenized by the text encoder and subsequently manifested in the image via joint attention.

MM-DiT computes the attention map as
$\mathrm{softmax} (  QK^{\top} /{\sqrt{d}})$,
where $Q = \mathrm{concat}(Q_{\mathrm{text}}, Q_{\mathrm{img}})$ and $K = \mathrm{concat}(K_{\mathrm{text}}, K_{\mathrm{img}})$
are the concatenated query and key matrices from text and image tokens, respectively. In this work, we focus on the cross-modal attention map defined as $A = \mathrm{softmax}(Q_{\rm img}\,K_{\rm text}^\top{/\sqrt d})$. 
We denote $A_{\tau}$ as the column of $A$ corresponding to text token $\tau$.

\begin{figure}[t!]
    \centering
    \includegraphics[width=\linewidth]{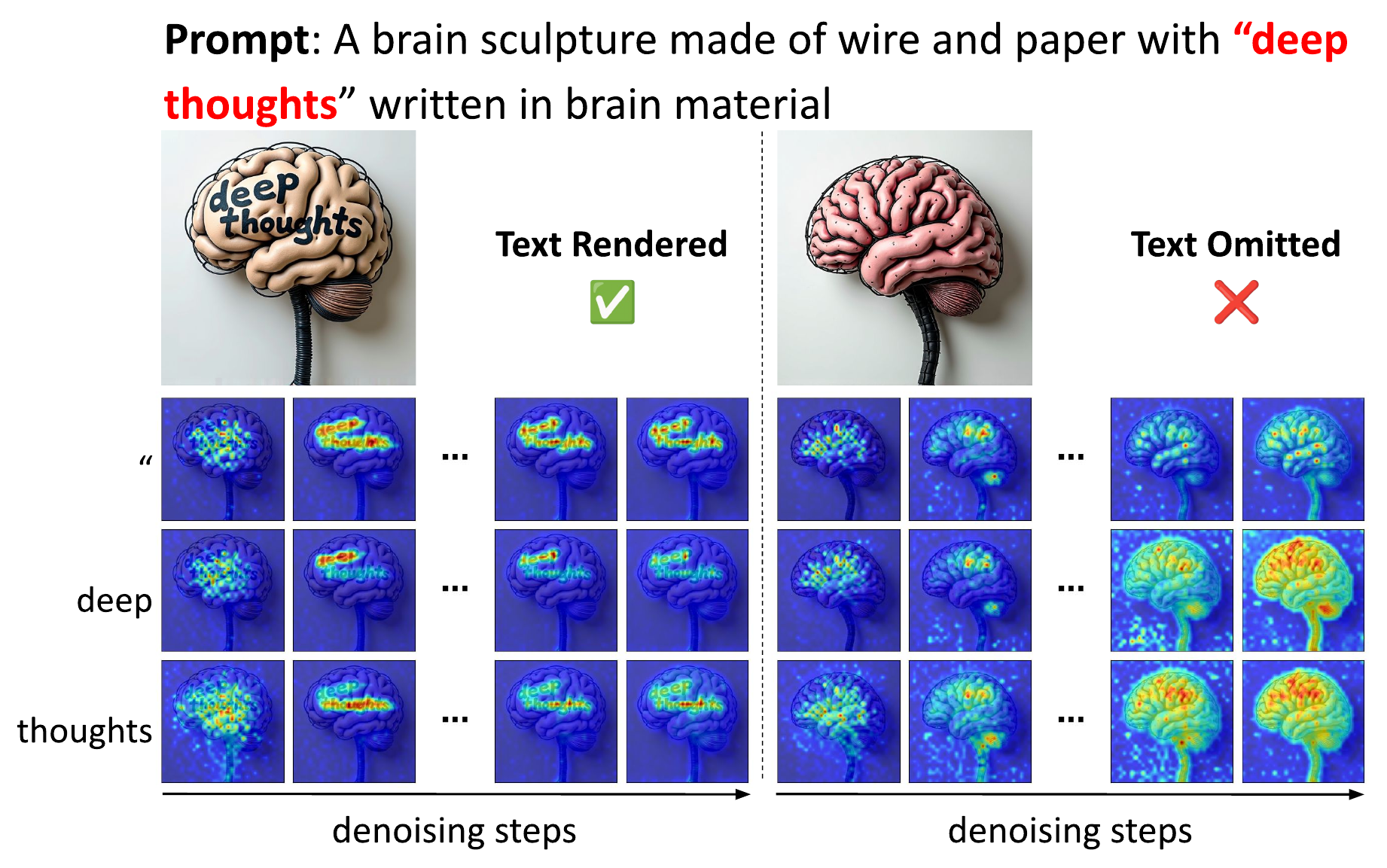}
    \caption{\textbf{Evolution of text token attention maps for successful vs. failed text rendering.} The left side shows a case where the text is successfully rendered, while the right side shows a failure case.  
Each row corresponds to the opening quotation mark token and each tokens in the phrase \textit{``deep thoughts''}. Attention maps are visualized across denoising steps, showing how the model attends to the spatial region associated with each token. 
Successful rendering is accompanied by strong, spatially aligned attention during the early timesteps, while unsuccessful cases exhibit weak or dispersed attention activations.  
The final generated images are shown in the upper part of the figure.}
    \label{fig:success_fail_attn_map}
\end{figure}

\paragraph{AMO sampler}

A flow-based generative model samples from a target distribution $\mathbf{Z}_1$ starting from a Gaussian noise distribution $\mathbf{Z}_0$ by solving the following discretized ODE using the Euler method:
\begin{equation}
\label{eq:euler_method}
\mathbf{Z}_{t_{k+1}} = \mathbf{Z}_{t_k} + \epsilon v_\theta(\mathbf{Z}_{t_k}, t_k),
\end{equation}
where $\epsilon = t_{k+1} - t_k$, $v_\theta$ is a velocity field parameterized by a neural network, and $t_0 = 0 < t_1 < \cdots < t_K = 1$ denotes the sequence of discretized timesteps.

To improve text rendering, AMO Sampler~\cite{hu2025amosamplerenhancingtext} augments Eq.~\ref{eq:euler_method} with an overshooting term scaled by a text-derived attention mask. At each step, the update rule is given by:

\small
\begin{align}
\mathbf{Z}_{t_{k+1}} 
&= \frac{t_{k+1}}{\mathbf{o}} \odot \Big( 
    \mathbf{Z}_{t_k} 
    + \epsilon(1 + c\mathbf{m}) \odot v_\theta(\mathbf{Z}_{t_k}, t_k) 
\Big) \nonumber \\
&\quad + \sqrt{(1 - t_{k+1})^2 - t_{k+1}^2 \frac{(1 - \mathbf{o})^2}{\mathbf{o}^2}} \odot \boldsymbol{\xi}
\end{align}

where $\boldsymbol{\xi} \sim \mathcal{N}(0, \mathbf{I})$, $\mathbf{o} = t_{k+1} + \epsilon c\mathbf{m}$, $c$ is a hyperparameter, and $\mathbf{m}$ and $\odot$ denote the attention-derived mask and element-wise product, respectively. In AMO Sampler, this mask is computed from the cross-modal attention map $\mathrm{softmax}(Q_{\text{text}} K_{\text{img}}^\top /{\sqrt{d}})$, where $Q_{\text{text}}$ is computed only from tokens corresponding to the textual content. We refer to this reversed-direction attention map as $A^{\mathrm{rev}}$, to distinguish it from our own formulation, which uses image tokens as queries and text tokens as keys instead. This targeted overshoot improves text rendering quality within the masked region without requiring full model fine-tuning, but its effectiveness depends on well-aligned attention maps.

\subsection{Attention Alignment and Text Omission}
\label{sec:analysis}

TextCrafter~\cite{du2025textcrafter} observes that the opening quotation mark token activates over the region where the text is rendered.
Building on this, we further investigate the relationship between the opening quotation mark token and the textual content tokens, comparing cases where the text is successfully rendered and when it is omitted (Figure~\ref{fig:success_fail_attn_map}).
Let $\boldsymbol{\tau}_{\text{text}} = \{\tau_1, \tau_2, ..., \tau_n\}$ be the textual content tokens to be generated in the image, and let $\tau_{\text{quo}}$ denote the opening quotation mark token. For example, given the prompt \textit{`A brain sculpture made of wire and paper with ``deep thoughts'' written in brain material'}, the textual content tokens $\boldsymbol{\tau}_{\text{text}}$ correspond to \textit{deep} and \textit{thoughts}, while the opening quotation mark token $\tau_{\text{quo}}$ corresponds to \textit{``}.

We observe that attention alignment in the early timesteps is strongly correlated with text omission.
In successful renderings (left of Figure~\ref{fig:success_fail_attn_map}), during the early timesteps, Flux’s cross-modal attention map $A_{\tau_{\text{quo}}}$ activates over the whole region where text is expected, effectively guiding the overall layout, while each $A_{\boldsymbol{\tau}_{\mathrm{text}}}$ focuses tightly on its own area. This early-stage separation and coverage in the attention maps help guide the model to focus consistently on the text throughout the generation process, ultimately leading to successful text rendering.
In contrast, in failure cases (right of Figure~\ref{fig:success_fail_attn_map}), both $A_{\tau_{\mathrm{quo}}}$ and $A_{\boldsymbol{\tau}_{\mathrm{text}}}$ show weak or misplaced activation, and attention drifts toward unrelated image elements, such as a brain‐shaped sculpture, resulting in text omission.

\begin{figure}[t!]
    \centering
    \includegraphics[width=\linewidth]{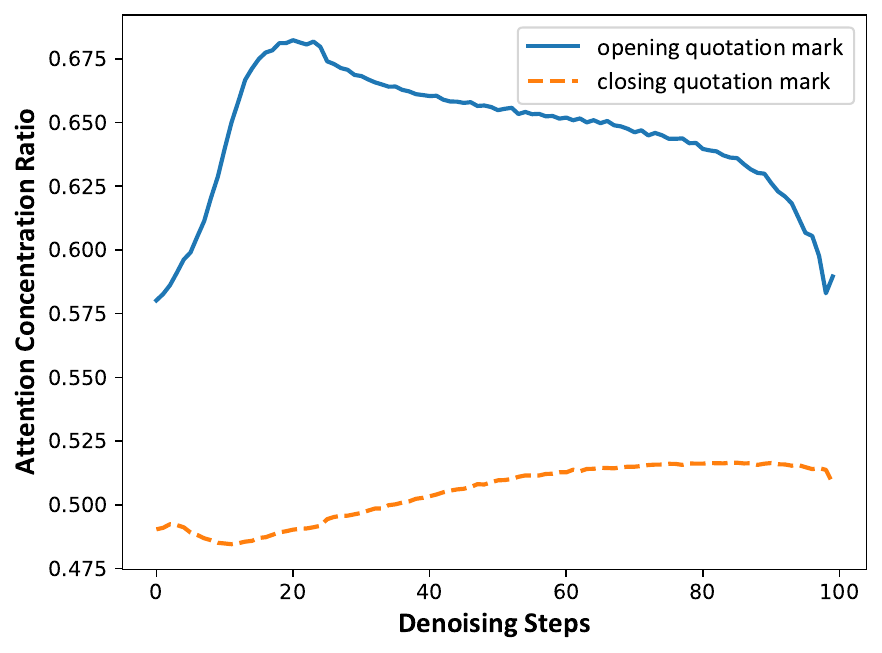}
    \caption{\textbf{Quantitative analysis of attention alignment with the ground-truth text region.} We visualize the Attention Concentration Ratio, defined as the ratio of the mean attention activation within the ground-truth text bounding box to the global mean activation for each denoising steps. The opening quotation mark (blue) exhibits a distinct peak during the early denoising stage (around $t=20$), followed by a decrease. In contrast, the closing quotation mark (orange) maintains significantly lower and flatter activation throughout the process.}
    \label{fig:quantitative_attn}
\end{figure}

To quantitatively validate our observation that $A_{\tau_{\text{quo}}}$ activates over the entire text region, we conduct the following experiment. We conduct an analysis using approximately 100 sampled images generated by Flux where the target text is correctly rendered. During the generation process, we track the cross-modal attention maps of the $\tau_{\text{quo}}$, using the closing quotation mark token as a baseline for comparison. For each timestep, we compute an attention concentration ratio, defined as the ratio of the mean attention activation within the text bounding box (detected via OCR~\cite{ppocr}) to the global mean attention activation across the entire image. We then average these scores across all samples for each denoising step. As shown in Figure~\ref{fig:quantitative_attn}, the opening quotation mark exhibits a distinct peak in attention concentration within the text region during the early denoising stage, followed by a decrease, whereas the closing quotation mark shows significantly lower and flatter activation. This confirms that the opening quotation mark plays a dominant role in establishing the text layout.

These findings indicate that precise alignment of the attention maps of the quotation mark token and textual content tokens with the local text regions during early denoising steps is critical for preventing text omission. In the next section, we introduce a test-time latent guidance strategy that enforces this alignment using two complementary loss terms.

\begin{figure*}[t!]
    \centering
    \includegraphics[width=\linewidth]{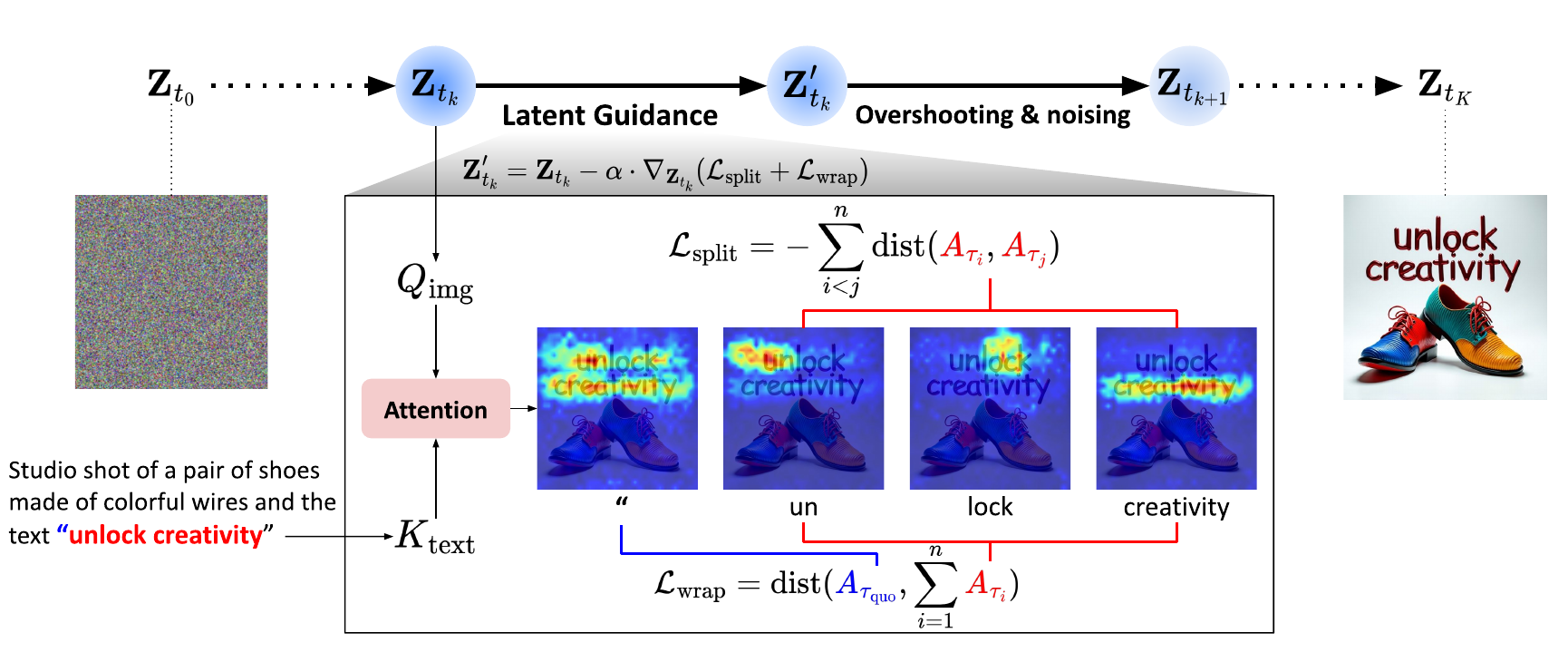}
    \caption{\textbf{Overview of \boldmethod{}.} In the early timesteps, we guide the latent before the denoising step, ensuring improved alignment between the textual contents and the text regions. This latent guidance is guided by gradients derived from the split and wrap losses, computed from attention maps corresponding to opening quotation marks and the desired text tokens.}
    \label{fig:method}
\end{figure*}

\subsection{Latent Guidance and Loss Function}
\label{sec:loss}
In this section, we introduce \boldmethod, a training-free latent guidance method. Before the denoising step, we refine the latent by guiding it based on the gradients from two proposed loss functions—the \textit{split loss} and \textit{wrap loss}. These losses are designed to enhance the alignment between textual content and corresponding text regions, leveraging the observations discussed in Section~\ref{sec:analysis}. We elaborate on these loss functions and their roles in detail below.

Prior methods~\cite{syngen, zhang2024object, li2023divide, meral2024conform} proposed various losses to align attention maps with the target task. In general, they encouraged the overlap of attention maps for semantically related tokens while minimizing the overlap for unrelated ones.
Our approach adapts this principle to text rendering. In this domain, however, it remains underexplored which groups of text tokens should share attention and which should be kept separate.
Based on our empirical findings, we design two complementary loss functions to reduce text omission via attention-based alignment.

Since textual content tokens rendered in the image should not overlap, we first ensure that $A_{\boldsymbol{\tau}_{\text{text}}}$ activates over spatially distinct regions corresponding to each token. We promote this separation using the split loss:

\begin{equation}
    \mathcal{L}_{\mathrm{split}} = - \sum_{i<j}^{n} \mathrm{dist}(A_{\tau_i}, A_{\tau_j}).
\end{equation}

This prevents overlap between attention regions for different textual content tokens by explicitly encouraging spatially distinct activations within the attention maps.
However, even when $A_{\boldsymbol{\tau}_{\text{text}}}$ is clearly separated, the target text does not always appear clearly in the image, as its attention may become aligned with unintended visual elements, causing the text to be visually overshadowed. We attribute this to cases where $A_{\tau_{\text{quo}}}$ fails to activate over the text regions, or where individual text tokens fall outside the area attended by $A_{\tau_{\text{quo}}}$. To address this, we propose the wrap loss:

\begin{equation}
    \mathcal{L}_{\mathrm{wrap}} = \mathrm{dist}(A_{\tau_{\text{quo}}}, \sum_{i=1}^n(A_{\tau_{\text{i}}})).
\end{equation}

This loss encourages the combined activated regions of $A_{\boldsymbol{\tau}_{\text{text}}}$ to overlap with the region activated by $A_{\tau_{\text{quo}}}$, thereby promoting more accurate text rendering in the image.
Following \cite{syngen}, we use a symmetric variant of the Kullback–Leibler (KL) divergence as a distance metric between attention maps, defined as

\small
\begin{align}
\mathrm{dist}(A_{\tau_i}, A_{\tau_j}) &= 
\frac{1}{2} D_{\mathrm{KL}}\left( 
    \frac{A_{\tau_i}}{\lVert A_{\tau_i} \rVert_1} 
    \,\middle\|\, 
    \frac{A_{\tau_j}}{\lVert A_{\tau_j} \rVert_1} 
\right) \nonumber \\
&\quad + 
\frac{1}{2} D_{\mathrm{KL}}\left( 
    \frac{A_{\tau_j}}{\lVert A_{\tau_j} \rVert_1} 
    \,\middle\|\, 
    \frac{A_{\tau_i}}{\lVert A_{\tau_i} \rVert_1} 
\right).
\end{align}

To ensure valid probability distributions, all attention maps used as inputs to the distance function are normalized to have a sum of 1.
This symmetric KL divergence serves as a flexible measure of similarity between attention maps, encouraging them to either converge on similar regions or remain distinct, depending on how the loss function is applied.
We define the final loss as 
\begin{equation}
\mathcal{L}=\frac{1}{N}\big(\mathcal{L}_{\mathrm{split}}+\mathcal{L}_{\mathrm{wrap}}\big),
\quad \text{where } N=\binom{n}{2}+1 .
\end{equation}
The normalization by $N$ accounts for the number of attention map comparisons used in the objective. 
Let $n = |\boldsymbol{\tau}_{\text{text}}|$ denote the number of textual content tokens. 
The split loss compares all unordered token pairs ($\binom{n}{2}$), and the wrap loss contributes one additional comparison.

Our full pipeline, illustrated in Figure~\ref{fig:method}, integrates the above losses for training-free latent guidance alongside the AMO sampler. 
At an early stage of the denoising step, we update the latent by $\mathbf{Z}_{t_k}'=\mathbf{Z}_{t_k}-\alpha\,\nabla_{\mathbf{Z}_{t_k}}\mathcal{L}$, 
where $\alpha$ is the guidance step size. We apply this guidance for the initial $t_\mathrm{guide}$ timesteps.
Since this latent guidance may push ${\mathbf{Z}}_{t_k}$ away from the model’s learned data manifold, we subsequently apply the AMO Sampler to correct this shift and guide the latent back toward a valid generative path.
To align our latent guidance with the editing process of the AMO Sampler, we use the attention map $A$ instead of $A^{\mathrm{rev}}$ as the spatial mask for AMO. We provide a detailed comparison of the two attention maps in Section~A.2 of the Supplementary Material.

%% file: sec/4_experiment.tex
\section{Experiment}
\label{experiment}

\begin{figure*}[t!]
    \centering
    \includegraphics[width=\linewidth]{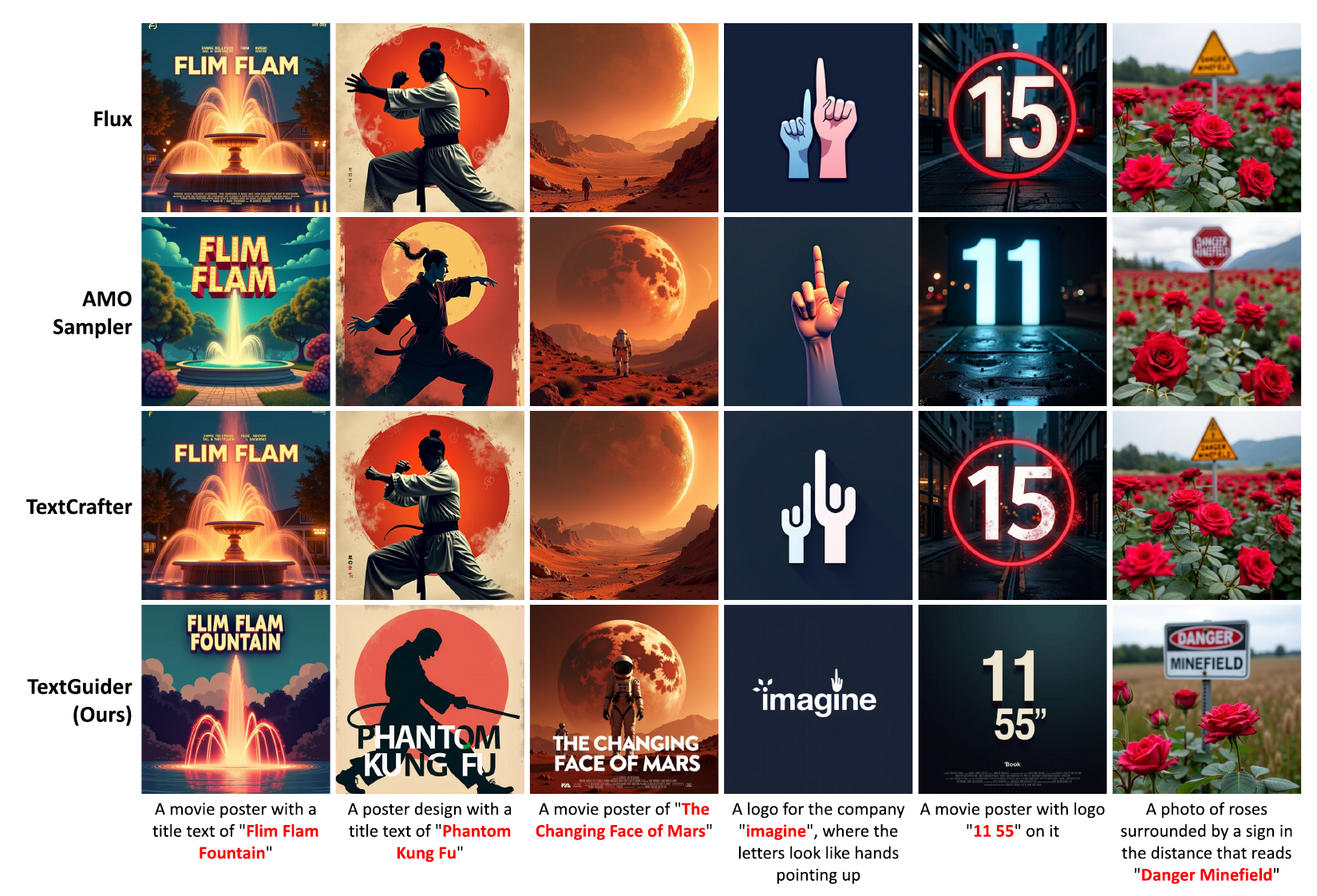}
    \caption{\textbf{Qualitative comparison of \boldmethod{} against baseline methods on the standard-text datasets.}} All generations are produced with the same random seed for consistency.
    \label{fig:main_figure}
\end{figure*}

\subsection{Experimental Settings}
\label{sec:experimental_settings}

We compare our method with three baselines: Flux~\cite{flux1-dev}, AMO Sampler~\cite{hu2025amosamplerenhancingtext}, and TextCrafter~\cite{du2025textcrafter}, the latter two of which are built on top of Flux. 
We apply our latent update strategy for the initial $t_\mathrm{guide}=25$ steps (the first quarter of the total 100 denoising steps) with a guidance step size $\alpha=60$.
All images are generated at a resolution of $512\times512$ pixels. We use 100 denoising steps for all methods except TextCrafter, which follows its original default setting of 50 steps for fair comparison. Additional implementation details are provided in Section~B of the Supplementary Material.

We construct our benchmark from three publicly available datasets: 
DrawTextCreative~\cite{liu2022character}, ChineseDrawText~\cite{ma2023glyphdraw}, and TMDBEval500~\cite{chen2023textdiffuser}.
We regard these as standard-text datasets: the target text averages about 15 characters and ranges from a single word to a sentence.
To further explore \boldmethod{}’s capability on long-text datasets, we additionally evaluate specific subsets of TextAtlasEval~\cite{textatlaseval}, namely textsceneshq, and styledtextsynth, selecting prompts whose target text length is 50–150 characters.

We evaluate the rendered text using PP-OCRv4~\cite{ppocr}, reporting Sentence Accuracy (Sen. Acc.) and Normalized Edit Distance (NED) as overall text rendering quality metrics, 
and word-level precision and recall to measure the correctness and completeness of the generated words. For the long-text datasets, we omit Sen.~Acc.\ because its exact match criterion becomes overly stringent for long strings. To assess the alignment between the prompt and the generated image, we also report the CLIP Score~\cite{clip}. 

We additionally conduct a human evaluation on the standard-text datasets to assess text accuracy and prompt alignment.
For text accuracy, we use the question "Which image has the highest text rendering accuracy?", and for prompt alignment, "Which image has the highest overall quality with respect to the given prompt?".
Participants rate each comparison using a three-way format: win, lose, or tie.
We evaluate 100 randomly selected prompts and collect 600 responses per comparison from a pool of 44 participants.

\begin{table}[t!]
\centering
\caption{\textbf{Text rendering performance on the standard-text datasets.} We compare our method with prior works using Sentence Accuracy, NED, Precision, Recall, and CLIP Score. All results are averaged over 3 runs with different noise seeds. The last row reports the mean score across all datasets, and the best score for each metric in each dataset is highlighted in bold.}
\label{tab:text_rendering}
\scriptsize  
\setlength{\tabcolsep}{1.5pt}  
\begin{tabular}{llccccc}
\toprule
\textbf{Dataset} & \textbf{Method} & \textbf{Sen. Acc.} & \textbf{NED} & \textbf{Precision} & \textbf{Recall} & \textbf{CLIP} \\
\midrule\midrule
\multirow{4}{*}{DrawTextCreative}
 & Flux~\cite{flux1-dev}         & 0.288 & 0.518 & 0.640 & 0.473 & 0.341 \\
 & AMO Sampler~\cite{hu2025amosamplerenhancingtext}  & 0.310 & 0.522 & 0.573 & 0.475 & 0.338 \\
 & TextCrafter~\cite{du2025textcrafter}  & 0.312 & 0.606 & 0.692 & 0.560 & \textbf{0.346} \\
 & \textbf{TextGuider} & \textbf{0.372} & \textbf{0.694} & \textbf{0.700} & \textbf{0.673} & 0.344 \\
\midrule
\multirow{4}{*}{ChineseDrawText}
 & Flux~\cite{flux1-dev}         & 0.278 & 0.587 & 0.593 & 0.533 & 0.337 \\
 & AMO Sampler~\cite{hu2025amosamplerenhancingtext}  & 0.261 & 0.567 & 0.583 & 0.527 & 0.335 \\
 & TextCrafter~\cite{du2025textcrafter}  & \textbf{0.341} & 0.673 & 0.639 & 0.615 & \textbf{0.343} \\
 & \textbf{TextGuider} & 0.339 & \textbf{0.693} & \textbf{0.664} & \textbf{0.660} & 0.340 \\
\midrule
\multirow{4}{*}{TMDBEval500}
 & Flux~\cite{flux1-dev}         & 0.358 & 0.644 & 0.665 & 0.551 & 0.345 \\
 & AMO Sampler~\cite{hu2025amosamplerenhancingtext}  & 0.383 & 0.629 & 0.733 & 0.526 & 0.338 \\
 & TextCrafter~\cite{du2025textcrafter}  & 0.410 & 0.739 & 0.692 & 0.621 & 0.355 \\
 & \textbf{TextGuider} & \textbf{0.479} & \textbf{0.837} & \textbf{0.745} & \textbf{0.724} & \textbf{0.361} \\
\midrule
\multirow{4}{*}{Mean}
 & Flux~\cite{flux1-dev}         & 0.326 & 0.607 & 0.639 & 0.532 & 0.342 \\
 & AMO Sampler~\cite{hu2025amosamplerenhancingtext}  & 0.339 & 0.594 & 0.654 & 0.517 & 0.337 \\
 & TextCrafter~\cite{du2025textcrafter}  & 0.375 & 0.698 & 0.676 & 0.608 & 0.350 \\
 & \textbf{TextGuider} & \textbf{0.425} & \textbf{0.775} & \textbf{0.713} & \textbf{0.697} & \textbf{0.353} \\
\bottomrule
\end{tabular}
\end{table}

\begin{table}[t]
\centering
\caption{\textbf{Text rendering performance averaged over long-text datasets.} We compare our method with prior works. The best score for each metric is highlighted in bold.}
\label{tab:long_text}
\scriptsize
\begin{tabular}{lcccc}
\toprule
\textbf{Method} & \textbf{NED} & \textbf{Precision} & \textbf{Recall} & \textbf{CLIP} \\
\midrule\midrule
  Flux~\cite{flux1-dev}          & 0.374 & 0.465 & 0.298 & 0.294 \\
AMO Sampler~\cite{hu2025amosamplerenhancingtext}   & 0.352 & 0.507 & 0.278 & 0.288 \\
\textbf{TextGuider}  & \textbf{0.546} & \textbf{0.614} & \textbf{0.433} & \textbf{0.321} \\
\bottomrule
\end{tabular}%
\end{table}

\subsection{Quantitative Results}
\label{sec:quantitative}

As shown in Table~\ref{tab:text_rendering}, our method outperforms all baselines on most metrics across the standard-text datasets (DrawTextCreative~\cite{liu2022character}, ChineseDrawText~\cite{ma2023glyphdraw}, and TMDBEval500~\cite{chen2023textdiffuser}).
Notably, our method achieves a substantial improvement in recall, indicating a reduction in text omission.
Averaged across datasets, our method improves recall by 31\% over Flux, 34\% over AMO Sampler, and 14\% over TextCrafter, with consistent gains across all datasets. We attribute this improvement to our approach’s enhancement of the alignment between textual tokens and image regions, which is especially beneficial for recovering omitted text.
In addition, our method improves overall text rendering quality, with notable gains in OCR-based Sen. Acc. and NED.
Our method also achieves high CLIP scores on average, indicating strong alignment between the generated images and the corresponding prompts.
Additional results for varying the number of denoising steps are provided in Section~C of the Supplementary Material.

AMO Sampler improves precision compared to Flux and enhances overall text rendering quality as reflected in higher Sen. Acc. Nevertheless, it has little impact on recall improvement.
TextCrafter, which generates text components separately, encourages the model to focus more on the text, leading to improved recall. However, its overall performance including recall remains inferior to ours, indicating the effectiveness of our attention-guided alignment strategy.

Table~\ref{tab:long_text} shows that, for long-text datasets, our method outperforms prior approaches across all metrics.
As the target text length increases, omissions become more likely; this is evident in the sharper decline in recall in the long-text setting (Table~\ref{tab:long_text}) compared to the standard-text datasets (Table~\ref{tab:text_rendering}). 
Even under this more challenging setting, our method improves recall by 45\% over Flux and 55\% over AMO Sampler, while also delivering substantial gains on other text-accuracy metrics and on the CLIP Score.



\begin{figure}[t]
    \centering
    \includegraphics[width=\columnwidth]{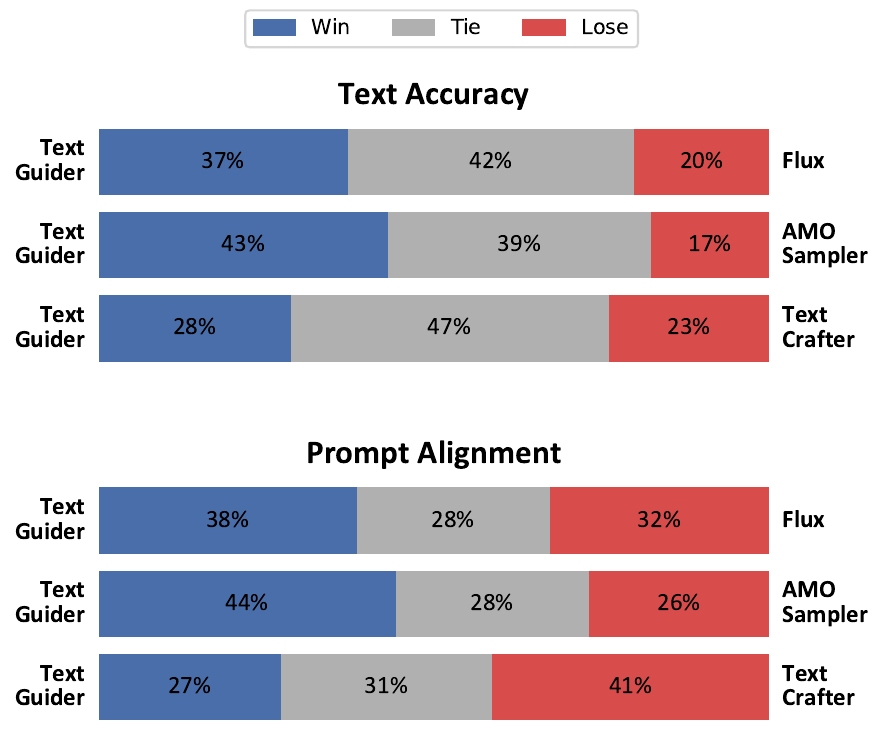}
    \caption{\textbf{Human evaluation results by metric.} Participants compare images based on text accuracy and prompt alignment using a three-way format (win/lose/tie). Evaluations use 100 prompts with 600 responses collected for each comparison setting.}
    \label{fig:human_eval}
\end{figure}

\paragraph{Human Evaluation}
To assess superiority in human perception, we conduct A/B preference tests comparing \boldmethod{} against each baseline (Figure~\ref{fig:human_eval}). Our method outperforms all other approaches in terms of text accuracy, reflecting its strength in faithfully rendering the textual content specified in the prompt.
Despite focusing solely on aligning text-related tokens with the correct image regions, our method maintains a competitive level of prompt alignment—comparable to Flux and better than AMO Sampler.

\subsection{Qualitative Results}
\label{sec:qualitative}
In Figure~\ref{fig:main_figure}, we compare the visual quality of training-free methods on the standard-text datasets.
In the first column, while other methods fail to generate the word ``Fountain'', our method successfully renders all words in the prompt. 
In the fifth column, Flux and TextCrafter incorrectly generate ``15'' by mixing ``11'' and ``55'', while AMO Sampler modifies ``15'' into ``11'' through overshooting, still omitting ``55''. In contrast, our method accurately generates all of the target words.

Across the remaining examples, while other methods fail to render any legible text, our method consistently produces the correct text. 
Our method not only excels at generating all text within the image but also faithfully adheres to the prompt.
For example, in the fourth column of Figure~\ref{fig:main_figure}, it accurately preserves the described text style,in which the letters are depicted as hands pointing upward.
For additional qualitative results, see Section E of the Supplementary Material.


Figure~\ref{fig:long_text} shows that text omission occurs more frequently in the long-text datasets. Other methods often fail to render the text correctly. For example, in the second and third rows, parts of the sentence are missing or appear blurred, whereas in the first and fourth rows the entire sentence is omitted. In contrast, our approach renders the text faithfully even for long sentences. Moreover, although the long-text prompts include more detailed specifications for visual elements, our method preserves these visual attributes without degradation (see Section I of the Supplementary Material for the prompts).

\begin{figure}[t]
    \centering
    \includegraphics[width=\linewidth]{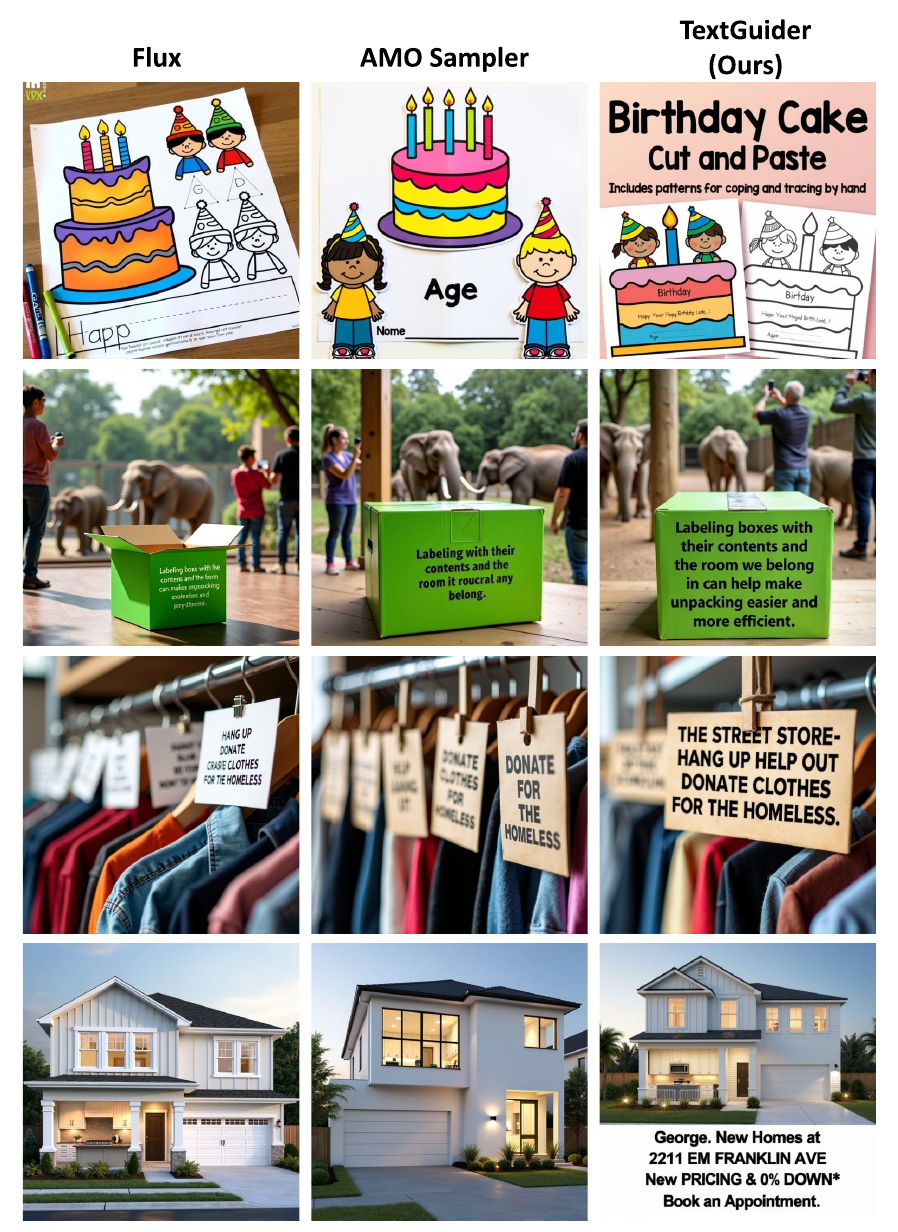}
    \caption{\textbf{Qualitative comparison of \boldmethod{} against baseline methods on the long-text datasets.} All generations are produced with the same random seed for consistency. Prompts are provided in Section I of the Supplementary Material.}
    \label{fig:long_text}
\end{figure}

\subsection{Ablation Study}
\label{sec:ablation}
We conduct an ablation study on the standard-text datasets to evaluate the contribution of each component (split loss, wrap loss, and AMO sampler) in our method. As shown in Table~\ref{tab:ablation_components}, using only the split loss yields a significant improvement in recall compared to using the AMO Sampler alone. While adding only the wrap loss leads to a drop in precision, it improves recall compared to using the AMO Sampler alone, resulting in a slight overall gain in text rendering quality (as measured by Sen. Acc. and NED).
Even without the AMO Sampler, using both losses yields substantial performance improvement.
However, the best results are achieved when all three components—AMO Sampler, split loss, and wrap loss—are combined.

\begin{figure}[t]
    \centering
    \includegraphics[width=\linewidth]{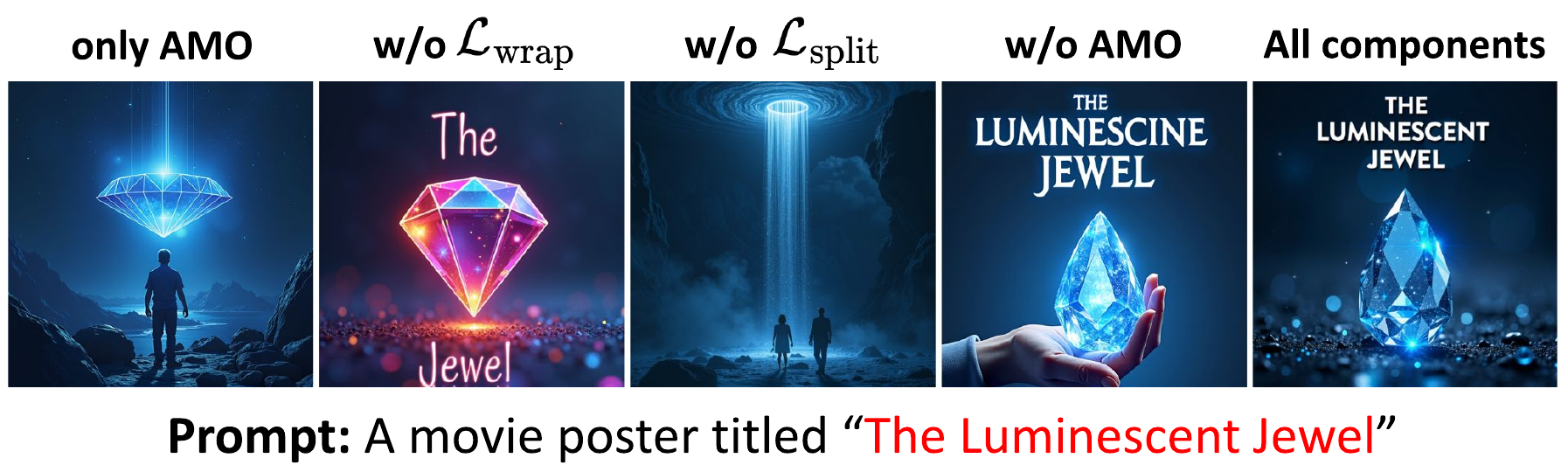}
    \caption{\textbf{Qualitative ablation study.} We compare the full method (rightmost) against four variants: only using the AMO Sampler, removing $\mathcal{L}_{\text{wrap}}$, removing $\mathcal{L}_{\text{split}}$, and removing the AMO Sampler entirely. Each variant shows a decline in text rendering quality, demonstrating the complementary roles of AMO and our proposed losses.}
    \label{fig:ablation}
\end{figure}

As illustrated in Figure~\ref{fig:ablation}, using only the AMO Sampler or excluding the split loss fails to generate the target text.
When the wrap loss is excluded, the phrase ``The Jewel'' appears correctly, but the word ``Luminescent'' is omitted.
Conversely, when the AMO Sampler is removed, all text appears, but with spelling errors.
Complete and accurate text rendering is achieved only when all components are applied together. Additional ablation studies are provided in Section D of the supplementary material.

%% file: sec/5_conclusion.tex
\section{Conclusion}
\label{sec:conclusion}


In this work, we propose {\boldmethod}, a method that addresses text omission through training-free latent guidance. Specifically, it guides the attention map of the textual contents to be aligned with the appropriate text regions.
We observe that text-related tokens in the prompt carry spatial information through the cross-attention map, especially in the early stages of the denoising steps.
In particular, the quotation mark token tends to activate over the entire region where text should appear, while each textual content token activates more locally over its corresponding area.
Building on this observation, we introduce a latent guidance strategy that encourages the model to render text more completely.
Specifically, we design a split loss, which separates the activations of textual content tokens, and a wrap loss, which ensures the quotation mark token attends to the entire text region.
Experimental results show that our method achieves state-of-the-art performance in test-time text rendering, demonstrating its ability to activate all relevant text tokens and effectively preventing text omission.

\begin{figure}[t]
    \centering
    \includegraphics[width=\linewidth]{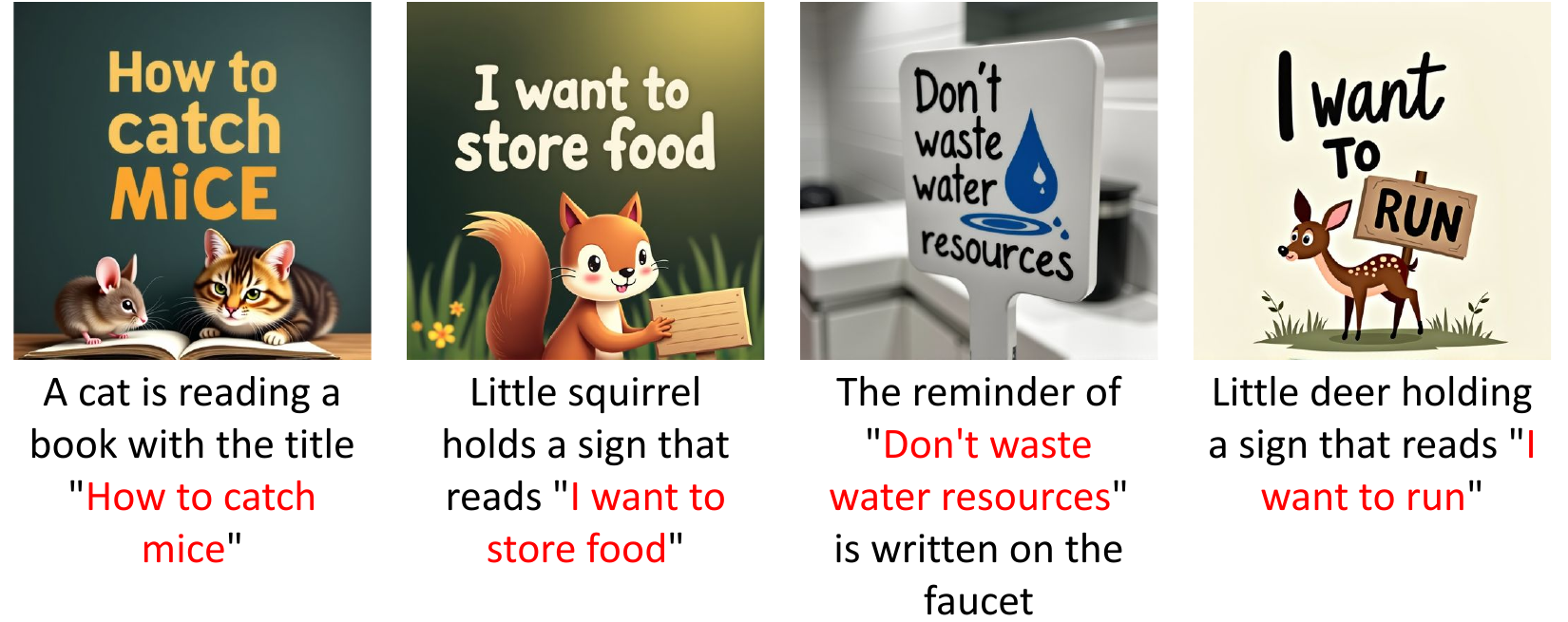}
    \caption{
\textbf{Examples illustrating the limitation of spatial control in text placement.}
While the generated text matches the prompt semantically, it may sometimes deviate from the spatial alignment specified by the prompt.
}
    \label{fig:limitation}
\end{figure}

While our method improves text rendering by promoting alignment between textual tokens and attended regions, it does not offer fine-grained spatial control over where the text appears within the image.
As illustrated in Figure~\ref{fig:limitation}, although the generated text content generally matches the prompt, its spatial placement is often misaligned with the location specified in the prompt.
For example, in the first image, the phrase \textit{``How to catch mice''} appears above the book rather than on the cover, and in other examples, the text is misaligned with signs or objects it is intended to appear.
These examples indicate a limitation of our current approach, which focuses on \emph{what} text is rendered rather than \emph{where} it should appear in the scene.
We leave explicit spatial control of text placement as a promising direction for future work.

\begin{table}[t] 
  \centering
  \caption{\textbf{Ablation studies over $t_\mathrm{guide}$ and $\alpha$.} The best score for each metric is in bold.}
  \label{tab:ablation_hyperparameter}
  
  \begin{subtable}{\linewidth} 
    \centering
    \caption{Ablation over $t_\mathrm{guide}$ (with $\alpha$ fixed at 60).}
    \label{tab:ablation_t_guide}
    \scriptsize
      \begin{tabular}{lccccc}
        \toprule
        \textbf{$t_\mathrm{guide}$} &  \textbf{Sen. Acc.} & \textbf{NED} & \textbf{Precision} & \textbf{Recall} & \textbf{CLIP} \\
        \midrule\midrule
        0(=AMO) & 0.252 & 0.541 & 0.634 & 0.483 & 0.342 \\
        10 & 0.310 & 0.572 & \textbf{0.702} & 0.547 & 0.344 \\
        \textbf{25(=Ours)}    & \textbf{0.342} & 0.656 & 0.700 & 0.635 & 0.345 \\
        50 & 0.265 & \textbf{0.661} & 0.620 & \textbf{0.654} & \textbf{0.349} \\
        75 & 0.206 & 0.629 & 0.564 & 0.594 & \textbf{0.349} \\
        \bottomrule
      \end{tabular}%
  \end{subtable}
  
  \vspace{1em} 

  \begin{subtable}{\linewidth} 
    \centering
    \caption{Ablation over $\alpha$ (with $t_\mathrm{guide}$ fixed at 25).}
    \label{tab:ablation_alpha}
    \scriptsize
      \begin{tabular}{lccccc}
        \toprule
        \textbf{$\alpha$} & \textbf{Sen. Acc.} & \textbf{NED} & \textbf{Precision} & \textbf{Recall} & \textbf{CLIP} \\
        \midrule\midrule
        0(=AMO) & 0.252 & 0.541 & 0.634 & 0.483 & 0.342 \\
        20 & 0.323 & 0.608 & 0.531 & 0.586 & 0.342 \\
        40 & 0.329 & 0.623 & 0.523 & 0.603 & 0.346 \\
        \textbf{60(=Ours)}    & \textbf{0.342} & 0.656 & \textbf{0.700} & 0.635 & 0.345 \\
        80 & 0.310 & \textbf{0.683} & 0.691 & 0.674 & \textbf{0.347} \\
        100 & 0.316 & 0.680 & 0.658 & \textbf{0.689} & 0.346 \\
        \bottomrule
      \end{tabular}%
  \end{subtable}

\end{table}

%% file: sec/a_appendix.tex
\setcounter{section}{0}
\renewcommand\thesection{\Alph{section}}

\maketitlesupplementary

\setcounter{table}{0}
\renewcommand{\thetable}{S\arabic{table}}
\setcounter{figure}{0}
\renewcommand{\thefigure}{S\arabic{figure}}
\setcounter{equation}{0}
\renewcommand{\theequation}{S\arabic{equation}}

\section{Additional Visualization}

\subsection{Effect of Loss Design on Attention Alignment}

\begin{figure}[htbp]
    \centering
    \includegraphics[width=\linewidth]{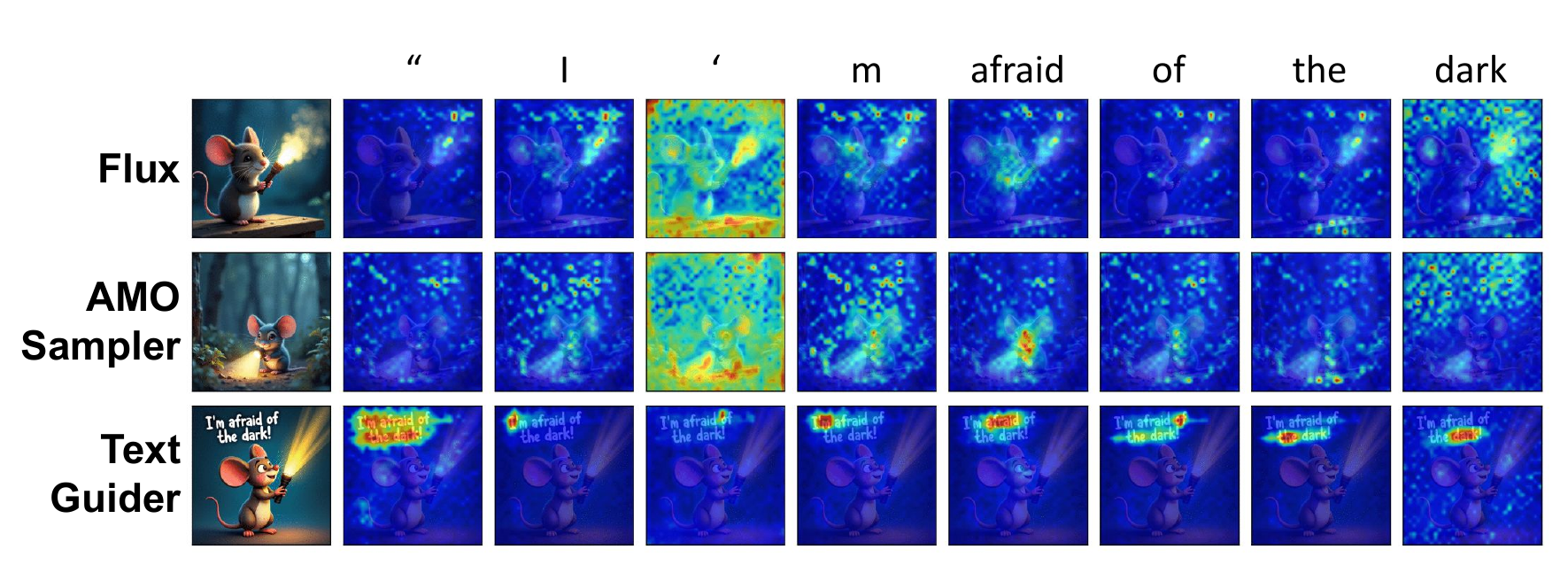}
    \caption{\textbf{Comparison of generated images and early-stage attention maps (extracted at 20 steps out of 100) for each method.} The prompt is `A mouse with a flashlight says ``\textit{I'm afraid of the dark}'' '. The first column shows the generated image, and the remaining columns visualize the attention maps corresponding to each text token. Rows correspond to different methods: Flux, AMO Sampler, and Ours.}
    \label{fig:comparison}
\end{figure}

As discussed in the main paper, we analyze the attention maps in cases where text is successfully rendered versus omitted, and guide the latent accordingly to follow the attention patterns observed in successful generations. In this section, we examine whether our proposed losses behave as intended by inspecting their effects on the resulting attention maps.
In Figure~\ref{fig:comparison}, we compare the attention maps across Flux, AMO Sampler, and our method.
Flux~\cite{flux1-dev} and AMO Sampler~\cite{hu2025amosamplerenhancingtext} often fail to produce meaningful activations, either lacking strong signals for key tokens (e.g., the quotation mark or “I”) or attending to irrelevant regions (e.g., “m” and “afraid” in AMO Sampler), which leads to text omission.
In contrast, our method yields semantically aligned attention: the quotation mark token activates broadly over the full text region, while each textual content token shows localized activation over its corresponding region—patterns that are consistently observed in successful generations by Flux.

\subsection{Comparison of Attention Maps Used in AMO Sampler and TextGuider}

\begin{figure}[htbp]
    \centering
    \includegraphics[width=\linewidth]{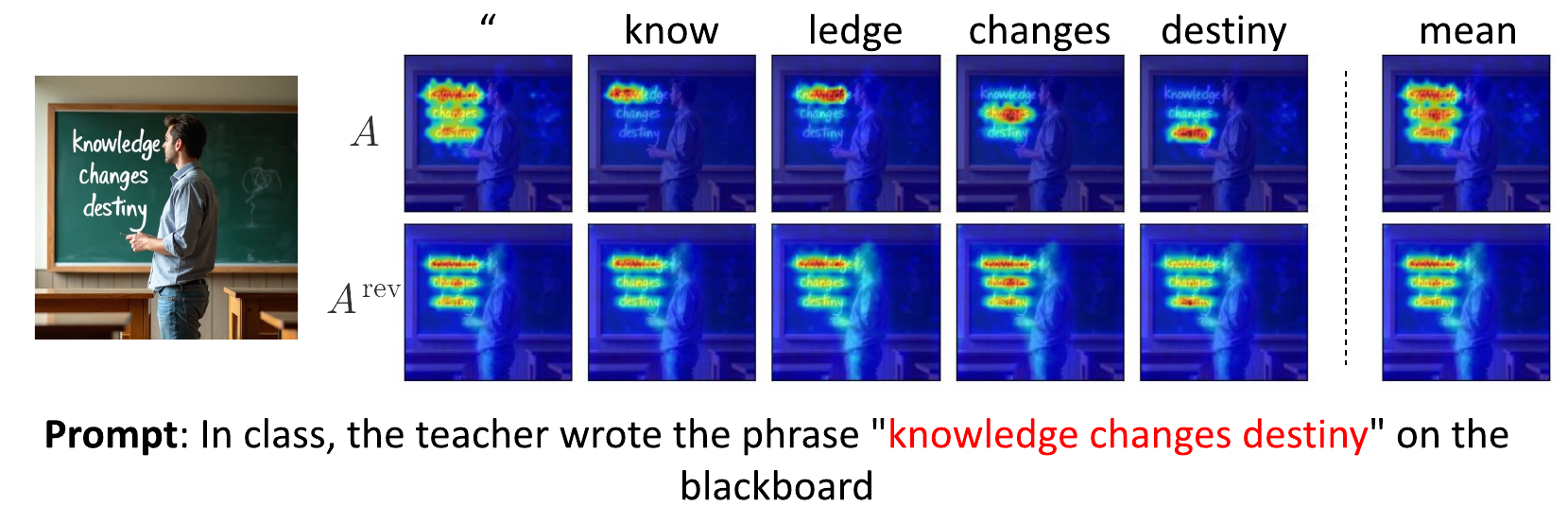}
    \caption{\textbf{Comparison of attention maps $A$ and $A^{\mathrm{rev}}$ for the phrase ``\textit{knowledge changes destiny}'' including the opening quotation mark.}
The leftmost image is the generated result from Flux used for attention analysis. Each subsequent column shows the attention map corresponding to a specific token. The final column shows the averaged attention map used as an overshooting mask.
}
    \label{fig:amo_ours_attn}
\end{figure}

Unlike AMO Sampler~\cite{hu2025amosamplerenhancingtext}, which uses the reverse attention map $A^{\mathrm{rev}}$, with the text as the query and the image as the key, our method uses the attention map $A$, with the image as the query and the text as the key.
We apply this attention map both for latent guidance and overshooting mask computation.
In Figure~\ref{fig:amo_ours_attn}, we compare $A$ and $A^{\mathrm{rev}}$ using an image generated by Flux~\cite{flux1-dev}.

For latent guidance, it is important that the attention maps for individual textual content tokens are spatially well-separated.
In the top row ($A$), we observe that each token (e.g., “know”, “changes”, “destiny”) activates a distinct and localized region on the blackboard where the corresponding word is rendered.
In contrast, in the bottom row ($A^{\mathrm{rev}}$), most tokens activate over the same broad region, making it difficult to isolate each token’s spatial focus.
This clear separation in $A$ supports our choice of using it for latent guidance.

For the overshooting mask, shown in the rightmost column, we take the average of the attention maps across all textual tokens.
In this case, both $A$ and $A^{\mathrm{rev}}$ attend to the overall text region on the blackboard, and their results appear visually similar.

We additionally compare the use of $A$ versus $A^{\mathrm{rev}}$ as the attention map for both the overshooting mask and latent guidance in Table~\ref{tab:attn_ablation}.
For AMO Sampler, there is no significant performance difference between using $A$ and $A^{\mathrm{rev}}$ for the overshooting mask.
However, for TextGuider, using $A$ leads to consistently better performance across all metrics compared to using $A^{\mathrm{rev}}$, demonstrating the effectiveness of $A$ for both latent guidance and overshooting.

\begin{table}[htbp]
\centering
\caption{\textbf{Ablation study on attention map choice ($A$ vs.\ $A^{\mathrm{rev}}$) for AMO Sampler~\cite{hu2025amosamplerenhancingtext} and TextGuider.} ($A$) and ($A^{\mathrm{rev}}$) denote the attention maps used for latent guidance and overshooting mask computation.}
\footnotesize  
\setlength{\tabcolsep}{3pt}  
\begin{tabular}{lccccc}
\toprule
\textbf{Method} & \textbf{Sen. Acc.} & \textbf{NED} & \textbf{Precision} & \textbf{Recall} & \textbf{CLIP} \\
\midrule\midrule
AMO Sampler ($A$)             & 0.340 & 0.596 & 0.636 & 0.513 & 0.336 \\
AMO Sampler ($A^{\mathrm{rev}}$) & 0.334 & 0.594 & 0.661 & 0.515 & 0.339 \\
\midrule
TextGuider ($A$)               & 0.424 & 0.773 & 0.707 & 0.696 & 0.354 \\
TextGuider ($A^{\mathrm{rev}}$) & 0.308 & 0.584 & 0.662 & 0.504 & 0.337 \\
\bottomrule

\end{tabular}
\label{tab:attn_ablation}
\end{table}

\section{Experimental Details}

We describe the details of gradient computation in Flux.
Flux consists of a combination of dual-stream and single-stream blocks.
We compute the latent gradients using all 19 layers of dual-stream blocks, in which the prompt and image are processed separately, while excluding the single-stream blocks from gradient computation.
For efficient memory management during backpropagation, we apply gradient checkpointing during computing latent gradients.

For other experimental settings, we follow standard configurations.
All experiments use a guidance scale of 5.0.
For other hyperparameters related to the AMO Sampler~\cite{hu2025amosamplerenhancingtext} and TextCrafter~\cite{du2025textcrafter}, we adopt the publicly released default settings.

All experiments are conducted using a single NVIDIA A40 GPU with 48GB of memory. Regarding GPU memory usage on this setup, the peak footprints are: model loaded—32{,}462~MiB; generating one image with Flux—33{,}066~MiB; generating one image with \textbf{TextGuider}—37{,}294~MiB.
In terms of runtime, generating a single image takes approximately 25 seconds with Flux and AMO Sampler, 32 seconds with TextCrafter, and 33 seconds with our method. For TextCrafter, the longer runtime is primarily due to the additional cost introduced by the pre-generation and region insulation steps, despite using only 50 denoising steps.
Our method also incurs additional overhead due to latent guidance.
However, as shown in Table~\ref{tab:timestep_ablation}, it achieves superior performance over both Flux and the AMO Sampler even with just 20 denoising steps, demonstrating the efficiency of our approach.



\section{Performance by Denoising Steps}
We evaluate performance across different timesteps, averaged over standard-text datasets (Table~\ref{tab:timestep_ablation}). Our method outperforms the baselines at all timestep settings. Moreover, even at just 20 steps, it outperforms both Flux and AMO Sampler at 100 steps in all metrics except for precision, demonstrating the effectiveness of early-stage guidance.

\begin{table}[t]
\centering
\caption{\textbf{Comparison across different numbers of sampling steps (mean over standard-text datasets)}. We report results for 20, 50, and 100 sampling steps, comparing TextGuider with Flux and AMO Sampler. The best score for each metric at each step count is highlighted in bold.}
\label{tab:timestep_ablation}
\footnotesize
\setlength{\tabcolsep}{2pt}
\begin{tabular}{clcccccc}
\toprule
\textbf{Steps} & \textbf{Method} & \textbf{Sen. Acc.} & \textbf{NED} & \textbf{Precision} & \textbf{Recall} & \textbf{CLIP} \\
\midrule\midrule
\multirow{3}{*}{20}
 & Flux~\cite{flux1-dev}         & 0.294 & 0.609 & 0.576 & 0.499 & 0.342 \\
 & AMO Sampler~\cite{hu2025amosamplerenhancingtext}  & 0.329 & 0.612 & 0.609 & 0.509 & 0.342 \\
 & \textbf{TextGuider}         & \textbf{0.385} & \textbf{0.690} & \textbf{0.638} & \textbf{0.600} & \textbf{0.349} \\
\midrule
\multirow{3}{*}{50}
 & Flux~\cite{flux1-dev}         & 0.312 & 0.608 & 0.597 & 0.526 & 0.342 \\
 & AMO Sampler~\cite{hu2025amosamplerenhancingtext}  & 0.336 & 0.610 & 0.667 & 0.525 & 0.340 \\
 & \textbf{TextGuider}         & \textbf{0.388} & \textbf{0.727} & \textbf{0.699} & \textbf{0.643} & \textbf{0.351} \\
\midrule
\multirow{3}{*}{100}
 & Flux~\cite{flux1-dev}         & 0.323 & 0.602 & 0.601 & 0.516 & 0.343 \\
 & AMO Sampler~\cite{hu2025amosamplerenhancingtext}  & 0.334 & 0.594 & 0.661 & 0.515 & 0.339 \\
 & \textbf{TextGuider}         & \textbf{0.424} & \textbf{0.773} & \textbf{0.707} & \textbf{0.696} & \textbf{0.354} \\
\bottomrule
\end{tabular}
\end{table}

\section{Hyperparameter Ablation Study}
We conduct ablations on DrawTextCreative~\cite{liu2022character} for the two hyperparameters $t_\mathrm{guide}$ and $\alpha$. As shown in Table~\ref{tab:ablation_hyperparameter} (a), with only 10 guided steps (10\% of the total 100 steps) we already surpass the AMO Sampler baseline, indicating the effectiveness of our method. On the other hand, using 50–75 guided steps tends to over-constrain the process and degrade text accuracy. Although tuning $\alpha$ can further improve performance even at higher $t_\mathrm{guide}$ steps, we adopt $t_\mathrm{guide}=25$ as the default in the main experiments to balance performance and computational cost.

As shown in Table~\ref{tab:ablation_hyperparameter} (b), across a wide $\alpha$ range, our method consistently surpasses AMO Sampler on Sen.Acc, NED, and recall, indicating little sensitivity to this hyperparameter. While increasing $\alpha$ can consistently improve recall, it may introduce trade-offs such as a decrease in Sen. Acc. or precision. We select $\alpha=60$ as it provides a good balance across metrics.

\begin{table}[t]
\centering
\caption{\textbf{Ablation study on guidance components.} We analyze the contribution of the AMO Sampler, $\mathcal{L}_{\mathrm{split}}$, and $\mathcal{L}_{\mathrm{wrap}}$. The best result for each metric is highlighted in bold.}
\label{tab:ablation_components}
\scriptsize  
\setlength{\tabcolsep}{4.5pt}  
    \begin{tabular}{ccc|ccccc}
    \toprule
    \textbf{AMO} & \textbf{$\mathcal{L}_{\mathrm{split}}$} & \textbf{$\mathcal{L}_{\mathrm{wrap}}$} & \textbf{Sen. Acc.} & \textbf{NED} & \textbf{Precision} & \textbf{Recall} & \textbf{CLIP} \\
    \midrule\midrule
    \checkmark & \xmark & \xmark & 0.340 & 0.596 & 0.636 & 0.513 & 0.336 \\
    \checkmark & \checkmark & \xmark & 0.399 & \textbf{0.774} & 0.687 & 0.686 & 0.352 \\
    \checkmark & \xmark & \checkmark & 0.365 & 0.604 & 0.567 & 0.532 & 0.336 \\
    \xmark & \checkmark & \checkmark & 0.370 & 0.749 & 0.674 & 0.683 & \textbf{0.354} \\
    \checkmark & \checkmark & \checkmark & \textbf{0.424} & 0.773 & \textbf{0.707} & \textbf{0.696} & \textbf{0.354} \\
    \bottomrule
    \end{tabular}
\end{table}

\section{Additional Qualitative Samples}

We present additional qualitative results across different datasets in Figures~\ref{fig:appendix_textcreative},~\ref{fig:appendix_chinese}, and~\ref{fig:appendix_tmdb}.
Compared to other baselines including Flux~\cite{flux1-dev}, AMO Sampler~\cite{hu2025amosamplerenhancingtext}, and TextCrafter~\cite{du2025textcrafter}, TextGuider more effectively addresses the issue of text omission.
Our method consistently renders more complete and accurate text, even in challenging cases such as long phrases or multi-word expressions.

\begin{table}[t]
\centering
\caption{\textbf{Cross-model results (SD~3.5 and AuraFlow)
averaged over standard-text datasets.} Each backbone is evaluated with three variants (Base, AMO Sampler, \boldmethod{}).
The best result per backbone is in bold.}
\label{tab:other_models}
\scriptsize
\setlength{\tabcolsep}{4pt}
\resizebox{\linewidth}{!}{%
\begin{tabular}{l l c c c c}
\toprule
\textbf{Backbone} & \textbf{Method} & \textbf{NED} & \textbf{Precision} & \textbf{Recall} & \textbf{CLIP} \\
\midrule\midrule
\multirow{3}{*}{SD~3.5~\cite{sd_3_5}}
  & Base & 0.638 & 0.360 & 0.539 & \textbf{0.368} \\
  & AMO Sampler~\cite{hu2025amosamplerenhancingtext} & 0.660 & 0.409 & 0.563 & 0.365 \\
  & \textbf{\boldmethod{}} & \textbf{0.678} & \textbf{0.431} & \textbf{0.591} & 0.363 \\
\midrule
\multirow{3}{*}{AuraFlow~\cite{auraflow}}
  & Base & 0.378 & 0.163 & 0.165 & 0.333 \\
  & AMO Sampler~\cite{hu2025amosamplerenhancingtext} & 0.412 & 0.231 & 0.215 & 0.329 \\
  & \textbf{\boldmethod{}} & \textbf{0.428} & \textbf{0.254} & \textbf{0.255} & \textbf{0.334} \\
\bottomrule
\end{tabular}%
}
\end{table}

\section{Applicability to Different Models}

We conduct additional experiments on Stable Diffusion 3.5 Medium (SD~3.5)~\cite{sd3,sd_3_5} and AuraFlow~\cite{auraflow}.
Following the main setting, we generate $512\times512$ images using 100 denoising steps.
For each backbone, we evaluate three variants across standard-text datasets: the base model, the model with the AMO Sampler, and our method (Table~\ref{tab:other_models}).
Across both SD~3.5 and AuraFlow, our method consistently surpasses the base and AMO Sampler on OCR-based metrics (NED, precision, recall) while maintaining competitive CLIP score. These results show that \boldmethod{} is not restricted to Flux and transfers effectively to other models.

\section{Additional Evaluations: Image Quality and Text-Image Alignment}
\label{app:quality_alignment}
To verify that our guidance does not degrade image quality, we additionally evaluate image-quality and prompt–image alignment metrics.
The results are shown in Table~\ref{tab:appendix_quality_alignment}.
\noindent\textbf{Image quality (Fréchet Inception Distance~\cite{fid}).} We report FID to assess the overall realism and fidelity of the generated images. A lower FID indicates better visual quality. TextGuider attains the lowest FID, showing that our guidance improves text rendering while preserving or improving image quality.
\noindent\textbf{Overall preferences (HPS v2~\cite{hpsv2}, PickScore~\cite{pickscore}, ImageReward~\cite{imagereward}).} These metrics aim to predict human preferences for text-to-image generations. TextGuider clearly surpasses prior methods on PickScore and ImageReward, while remaining competitive on HPS v2.
\noindent\textbf{Text–image alignment (VQAScore~\cite{vqascore}).} VQAScore leverages multimodal LLMs to provide a nuanced alignment signal. TextGuider achieves the highest VQAScore, confirming effective alignment of textual content to the prompt without degrading overall image quality.

\begin{table}[t] 
\centering
\caption{\textbf{Image quality and text–image alignment (mean over standard-text datasets).} Lower is better for FID; higher is better otherwise. Best per column in bold.}
\label{tab:appendix_quality_alignment}
\scriptsize
\setlength{\tabcolsep}{3pt}
\resizebox{\columnwidth}{!}{%
\begin{tabular}{lccccc}
\toprule
\textbf{Method} & \textbf{FID~\cite{fid}} $\downarrow$ & \textbf{HPS v2}~\cite{hpsv2} $\uparrow$ &
\textbf{PickScore}~\cite{pickscore} $\uparrow$ &
\textbf{ImageReward}~\cite{imagereward} $\uparrow$ &
\textbf{VQAScore}~\cite{vqascore} $\uparrow$ \\
\midrule\midrule
Flux~\cite{flux1-dev} & 127.3 & \textbf{0.279} & 0.337 & 0.766 & 0.759 \\
AMO Sampler~\cite{hu2025amosamplerenhancingtext} & 130.1 & 0.274 & 0.308 & 0.675 & 0.729 \\
\textbf{TextGuider (Ours)} & \textbf{126.3} & 0.277 & \textbf{0.356} & \textbf{0.815} & \textbf{0.791} \\
\bottomrule
\end{tabular}%
}
\end{table}

\section{Comparison with Training-based Methods}
\begin{figure*}[htbp]
    \centering
    \includegraphics[width=\linewidth]{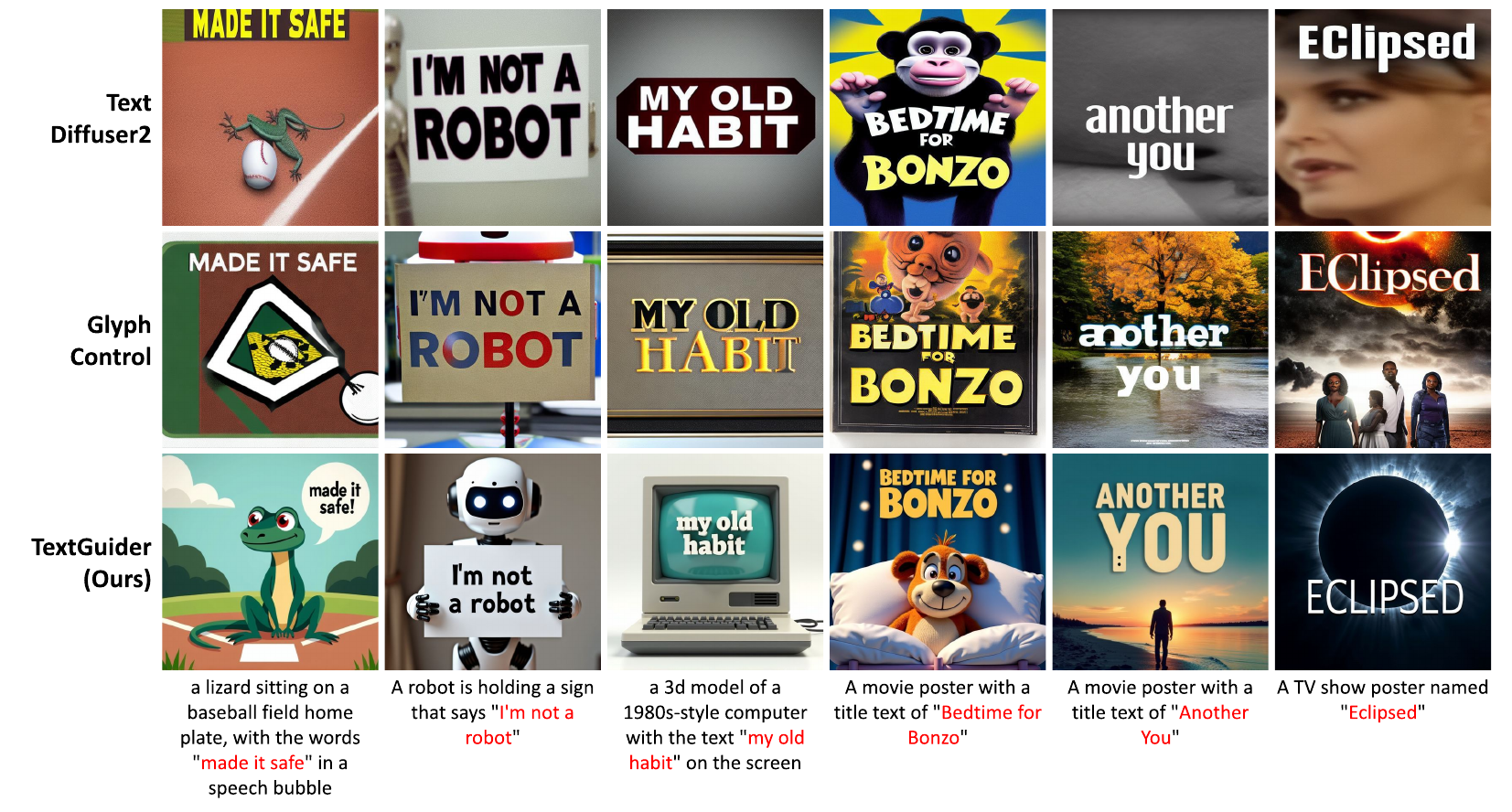}
    \caption{\textbf{Qualitative comparison with training-based methods.}
}
    \label{fig:comparison_training}
\end{figure*}

\begin{figure*}[htbp]
    \centering
    \includegraphics[width=\linewidth]{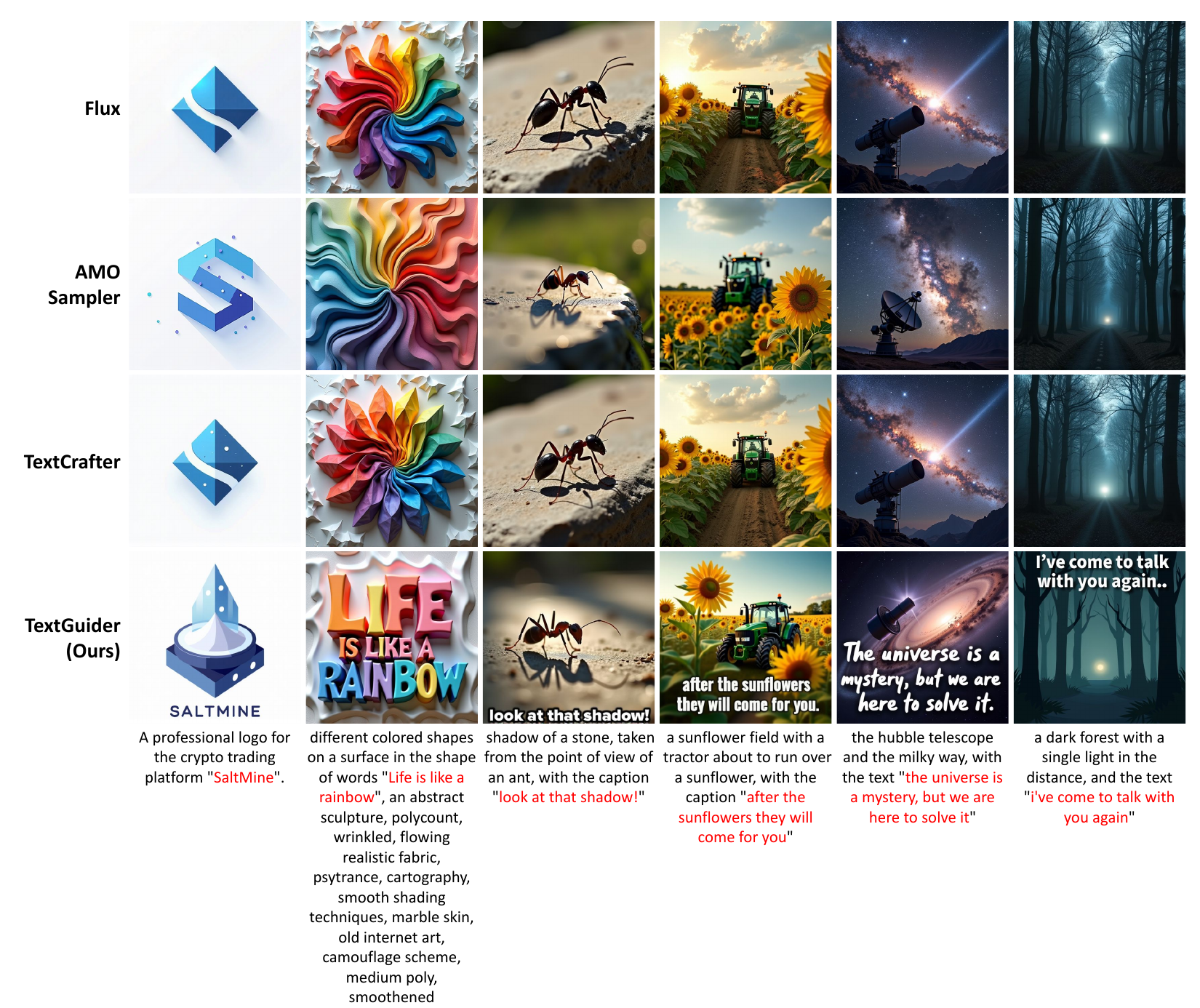}
    \caption{\textbf{Qualitative examples from the DrawTextCreative dataset~\cite{liu2022character}.}}
    \label{fig:appendix_textcreative}
\end{figure*}

\begin{figure*}[htbp]
    \centering
    \includegraphics[width=\linewidth]{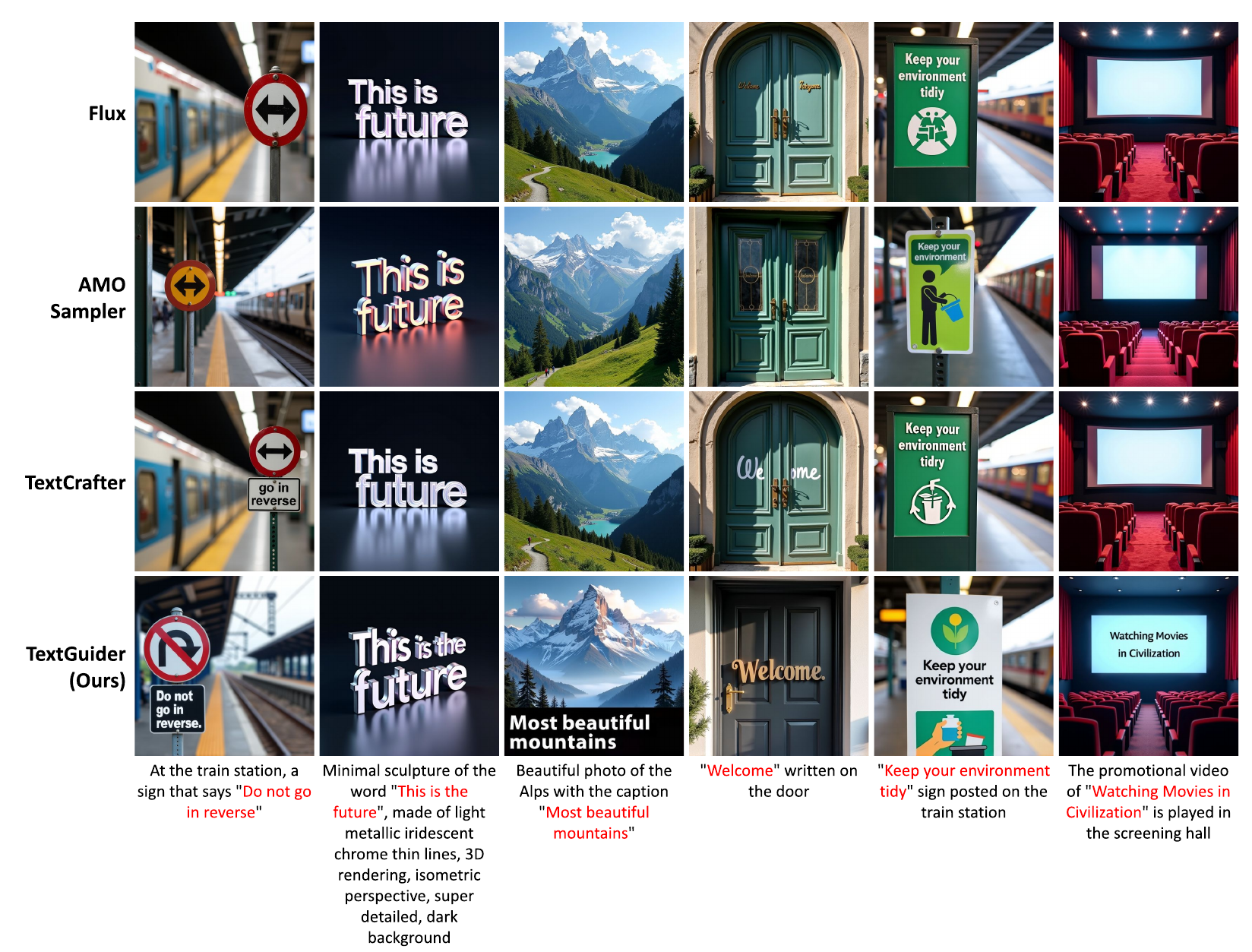}
    \caption{\textbf{Qualitative examples from the ChineseDrawText dataset~\cite{ma2023glyphdraw}.}}
    \label{fig:appendix_chinese}
\end{figure*}

\begin{figure*}[htbp]
    \centering
    \includegraphics[width=\linewidth]{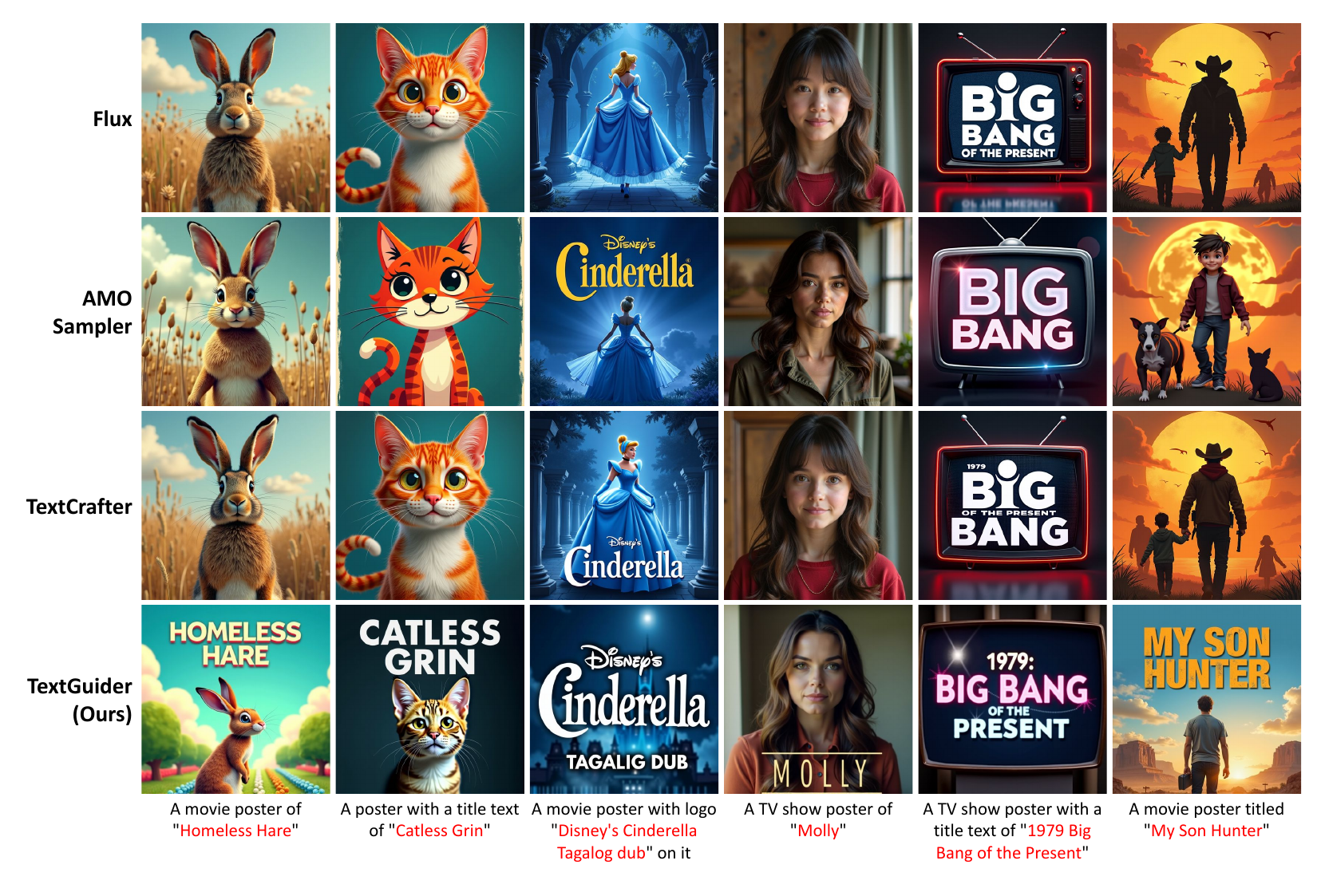}
    \caption{\textbf{Qualitative examples from the TMDBEval500 dataset~\cite{chen2023textdiffuser}.}}
    \label{fig:appendix_tmdb}
\end{figure*}

We further compare our method with training-based approaches (Table~\ref{tab:comparison_training}).  
Despite requiring no additional training, our method outperforms GlyphControl in all metrics except NED, and achieves higher recall compared to TextDiffuser2.  
Moreover, training-based methods tend to focus solely on text rendering, often failing to faithfully reflect other aspects of the prompt or maintain high visual quality (Figure~\ref{fig:comparison_training}).


\begin{table}[htbp]
\centering
\caption{\textbf{Comparison of text rendering performance between training-based and training-free methods (mean over standard-text datasets).}}
\scriptsize  
\setlength{\tabcolsep}{2pt}  
\begin{tabular}{llccccc}
\toprule
& Method & Sen. Acc. & NED & Precision & Recall & CLIP \\
\midrule\midrule
\multirow{2}{*}{\textbf{Training-based}} 
& TextDiffuser2~\cite{chen2024textdiffuser} & 0.464 & 0.852 & 0.792 & 0.651 & 0.349 \\
& GlyphControl~\cite{yang2023glyphcontrol}   & 0.328 & 0.819 & 0.604 & 0.487 & 0.351 \\
\midrule
\multirow{4}{*}{\textbf{Training-free}} 
& Flux~\cite{flux1-dev}          & 0.323 & 0.602 & 0.601 & 0.516 & 0.343 \\
& AMO Sampler~\cite{hu2025amosamplerenhancingtext}   & 0.334 & 0.594 & 0.661 & 0.515 & 0.339 \\
& TextCrafter~\cite{du2025textcrafter}   & 0.363 & 0.697 & 0.664 & 0.601 & 0.350 \\
& \textbf{TextGuider} & 0.424 & 0.773 & 0.707 & 0.696 & 0.354 \\
\bottomrule
\end{tabular}
\label{tab:comparison_training}
\end{table}

\section{Prompts for Long-Text Examples (Figure~5)}
Below we list the exact long-text prompts used for Figure~5 of the main paper.
The prompts are ordered to match the rows, from top to bottom.
\begin{enumerate}[label=Row~\arabic*:]
  \item We see a setting for a colorful birthday cake cut-and-paste activity featuring a layered cake with candles, two children in party hats, and space for writing age, labeled as ``\textit{Birthday Cake Cut and Paste Includes patterns For copying and tracing by hand.}''
  \item A zoo enclosure with lions and elephants, people taking photos, and a bright green packing box with text the text : ``\textit{Labeling boxes with their contents and the room they belong in can help make unpacking easier and more efficient}''
  \item Conveying a row of signs hangs above hangers with clothes, the annotation says: ``\textit{THE STREET STORE HANG UP HELP OUT DONATE CLOTHES FOR THE HOMELESS.}''
  \item We see a depiction of a modern two-story house with large windows and a garage. Inside, a bright kitchen features white cabinets, stainless steel appliances, and a large island, the text inside is: ``\textit{George. New Homes at 2211 EM FRANKLIN AVE New PRICING \& 0\% DOWN* Book an Appointment.}''
\end{enumerate}

